\documentclass[10pt]{book}

\usepackage[table]{xcolor}
\usepackage{mathrsfs} 
\usepackage{amssymb,amsmath}
\usepackage{graphicx}
\usepackage{hyperref} 
\usepackage{multirow}
\usepackage{xspace}
\usepackage{booktabs}
\usepackage{caption} 
\usepackage{longtable}

\usepackage{pdfpages}
\usepackage{caption}

\hypersetup{colorlinks,linkcolor={red!50!black},citecolor={blue!50!black},urlcolor={blue!80!black}}
\usepackage[top=1.2in, bottom=1.2in, left=1.4in, right=1.4in]{geometry}

\usepackage{tikz}
\usetikzlibrary{shapes,calc,positioning,automata,arrows,trees}

\newcommand\bcpen{\includegraphics[width=15pt]{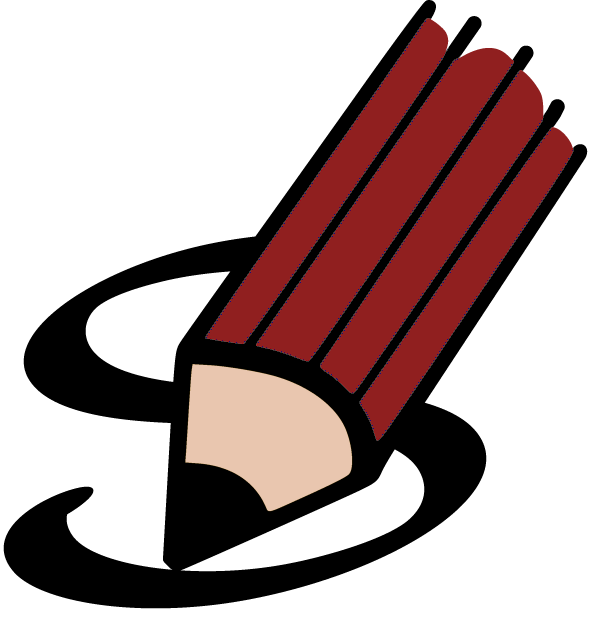}}

\usepackage{listingsutf8}
\usepackage{textcomp}
\usepackage{verbatim}

\definecolor{v3lgray}{gray}{0.98}
\definecolor{v2lgray}{gray}{0.85}
\definecolor{vlgray}{gray}{0.92}
\definecolor{dgray}{rgb}{0.4,0.4,0.4}
\definecolor{dblue}{RGB}{0,0,99}
\definecolor{dred}{RGB}{175,6,54}
\definecolor{dgreen}{RGB}{47,135,7}
\definecolor{dviolet}{RGB}{102,0,153}
\definecolor{mblue}{RGB}{0,0,180}
\definecolor{dorange}{RGB}{204, 82, 0}

\def\ace{ACE\xspace}

\def\x3{{\rm XCSP$^3$}\xspace}
\def\jv3{{\rm JvCSP$^3$}\xspace}
\def\p3{{\rm PyCSP$^3$}\xspace}

\newcommand{\gb}[1]{{\tt #1}} 

\lstdefinelanguage{json}{
    basewidth  = {.6em,0.6em},
    basicstyle=\normalfont\ttfamily,
    breaklines=true,
    morestring=[b]',
    morestring=[b]", 
    sensitive=false,
    stringstyle=\color[rgb]{0.227,0.226,0.441}\ttfamily, 
    escapechar=!,
    showstringspaces=false,
    xleftmargin=20pt, 
    breaklines=true,basicstyle=\ttfamily\small,inputencoding=utf8/latin9,texcl
}
\lstnewenvironment{json}{\lstset{language=json}}{}

\lstdefinelanguage{mcsp}{
  keywords={forall,array,block,class,implements,model,public,slide},
  basewidth  = {.6em,0.6em},
  keywordstyle=\color{dred}\bfseries,
  ndkeywords={intension,lessThan,lessEqual,greaterEqual,greaterThan,equal,different,implication,equivalence,conjunction,disjunction,extension,regular,mdd,allDifferent,allDifferentMatrix,allEqual,ordered,increasing,decreasing,strictlyIncreasing,strictlyDecreasing,lex,lexMatrix,sum,count,atMost,atLeast,exactly,atMost1,atleast1,exactly1,element,channel,maximum,minimum,cardinality,nValues,noOverlap,cumulative,instantiation,clause,circuit,minimize,maximize},
  ndkeywordstyle=\color{mblue}\bfseries,
  identifierstyle=\color{black},
  sensitive=false,
  comment=[l]{//},
  morecomment=[s]{<!--}{-->},
  commentstyle=\color{dgreen}\ttfamily,
  stringstyle=\color{dgreen}\rmfamily, 
  morestring=[b]',
  morestring=[b]",
  escapechar=~,
  showstringspaces=false,
  classoffset=2, morekeywords={private},keywordstyle=\color{gray},
  classoffset=3, morekeywords={dom,size,when},keywordstyle=\color{dorange},
  xleftmargin=-22pt,xrightmargin=-22pt,
  breaklines=true,basicstyle=\ttfamily\footnotesize,backgroundcolor=\color{v3lgray},inputencoding=utf8/latin9,texcl
}
\lstnewenvironment{mcsp}{\lstset{language=mcsp}}{}

\definecolor{colorex}{RGB}{255,248,220} 
\definecolor{mgray}{rgb}{0.55,0.55,0.55}
\definecolor{officegreen}{rgb}{0.0, 0.5, 0.0}

\newcounter{cntPy}

\lstnewenvironment{python}{\lstset{
    inputencoding=utf8/latin1,
    language=python,
    ndkeywords={size,dom,matrix,strict,occurrences},
    ndkeywordstyle=\color{mblue}, 
    keywords=[2]{data,satisfy,minimize,maximize,solve},
    keywordstyle=[2]\color{dred}, 
    deletekeywords={def},
    keywords=[3]{def,If,Then,Match},
    keywordstyle=[3]\color{dorange},
    commentstyle=\color{officegreen}, 
    showstringspaces=false,
    captionpos=t,
    basicstyle=\ttfamily\footnotesize,
    framesep=5pt,
    xleftmargin=5pt,xrightmargin=1pt,
    backgroundcolor=\color{colorex}, 
    escapechar=@
  }
}{}

\usepackage[skins,breakable,xparse]{tcolorbox}
\newcommand{\core}[1]{ 
  \medskip \begin{tcolorbox}[
    enhanced,breakable,
    boxsep=0pt,top=4pt,bottom=0pt,left=2mm,right=1mm,
    toprule=0.1mm,leftrule=0.1mm,rightrule=0.25mm,bottomrule=0.25mm,shadow={0.2mm}{-0.2mm}{0mm}{dgray},
    overlay unbroken and first={\node (logo) at ([xshift=4mm,yshift=-5mm]frame.north west) {#1}; },
    colframe=dgray,titlerule=-0.2mm,toptitle=3mm,coltitle=dred, fonttitle=\bfseries,
    lines before break=6, pad at break*=10pt
    }

\newenvironment{boxpy}
 {\stepcounter{cntPy} \core{\bcpen} , colback=colorex, title style={color=colorex}, title=~ ~ \,PyCSP$^3$ Model \thecntPy]}
 {\end{tcolorbox}}

\newenvironment{command}
  {\quote\small\verbatim}
  {\endverbatim\endquote}

  \usepackage{authblk}
  
 \title{\textcolor{dred}{Proceedings of the \x3 Competition 2022}}
\author{Gilles Audemard \and Christophe Lecoutre \and Emmanuel Lonca}

\affil{\vspace{1cm} CRIL \\University of Artois \& CNRS \\ France}

\date{September 1, 2022 \\ {\small (Revised$^*$ on December 10, 2023)}}

\begin{document}
\maketitle

~ \\
~ \\

\bigskip

This document represents the proceedings of the \x3 Competition 2022.
The website containing all {\bf detailed results} is available at:
  \href{https://www.cril.univ-artois.fr/XCSP22/}{https://www.cril.univ-artois.fr/XCSP22/}

  \bigskip
  \bigskip
\noindent The organization of this 2022 competition involved the following tasks:
\begin{itemize}
\item adjusting general details (dates, tracks, $\dots$) by G. Audemard, C. Lecoutre and E. Lonca
\item selecting instances (problems, models and data) by C. Lecoutre
\item receiving, testing and executing solvers on CRIL cluster by E. Lonca
\item validating solvers and rankings by C. Lecoutre and  E. Lonca
\item developping the 2022 website dedicated to results by G. Audemard
\end{itemize}

\bigskip\bigskip 
{\bf Important}: for reproducing the experiments and results, it is important to use the set of \x3 instances used in the competition.
These instances can be found in this \href{https://www.cril.univ-artois.fr/~lecoutre/compets/instancesXCSP22.zip}{archive}.
Some (usually minor) differences may exist when compiling the models presented in this document and those that can be found in this  \href{https://www.cril.univ-artois.fr/~lecoutre/compets/modelsXCSP22.zip}{archive}.

\bigskip
    {\bf Revision} ($^*$) of December 2023: some models in this document have been simplified while using new possibilities offered by Version 2.2 of \p3.
Note that in order to reproduce results and/or to make fair new comparisons with respect to solvers engaged in the 2022 competition, you have to use the very \href{https://www.cril.univ-artois.fr/~lecoutre/compets/instancesXCSP22.zip}{same set} of \x3 instances, as in the 2022 competition.

\tableofcontents

\chapter{About the Selection of Problems in 2022}

Remember that the complete description, {\bf Version 3.0.7}, of the format (\x3) used to represent combinatorial constrained problems can be found in \cite{BLAPxcsp3}.
For the 2022 competition, we have limited \x3 to its kernel, called \x3-core \cite{BLAP_xcsp3core}.
This means that the scope of \x3 is restricted to:
\begin{itemize}
\item integer variables,
\item CSP and COP problems, 
\item a set of 21 popular (global) constraints for Standard tracks:
  \begin{itemize}
  \item generic constraints: \gb{intension} and \gb{extension} (also called  \gb{table})
  \item language-based constraints: \gb{regular} and \gb{mdd}
  \item comparison constraints: \gb{allDifferent}, \gb{allEqual}, \gb{ordered} and \gb{lex}
  \item counting/summing constraints: \gb{sum}, \gb{count}, \gb{nValues} and \gb{cardinality}
  \item connection constraints: \gb{maximum}, \gb{minimum}, \gb{element} and \gb{channel}
  \item packing/scheduling constraints: \gb{noOverlap} and \gb{cumulative}
  \item \gb{circuit}, \gb{instantiation} and \gb{slide}
  \end{itemize}
  and a small set of constraints for Mini-solver tracks.
\end{itemize}

For the 2022 competition, 41 problems have been selected.
They are succinctly presented in Table \ref{fig:problems}.
For each problem, the type of optimization is indicated (if any), as well as the involved constraints.
At this point, do note that making a good selection of problems/instances is a difficult task.
In our opinion, important criteria for a good selection are:
\begin{itemize}
\item the novelty of problems, avoiding constraint solvers to overfit already published problems;
\item the diversity of constraints, trying to represent all of the most popular constraints (those from \x3-core) while paying attention to not over-representing some of them (in particular, second class citizens);
\item the scaling up of problems.
\end{itemize}

\begin{table}
  \begin{small}
  \begin{tabular}{p{4cm}p{3cm}p{7.5cm}}
    \toprule
CSP Problems & & Global Constraints\\
    \midrule
\rowcolor{vlgray}{}    Aztec Diamond & & \gb{table} ($*$) \\
    Blocked Queens &   & \gb{allDifferent} \\
\rowcolor{vlgray}{}    Car Sequencing &  & \gb{cardinality}, \gb{sum}, \gb{table} \\
    Costas Arrays & & \gb{allDifferent} \\
\rowcolor{vlgray}{}    Crosswords & & \gb{table} \\
    Crypto & & \gb{table} ($*$) \\
\rowcolor{vlgray}{}  Diamond Free & & \gb{lex}, \gb{sum}\\
Eternity & & \gb{allDifferent}, \gb{table} \\
\rowcolor{vlgray}{} Hadamard & & \gb{sum} \\
Hidato & & \gb{allDifferent}, \gb{table} ($*$) \\
\rowcolor{vlgray}{} Knight Tour & & \gb{circuit} \\
Molnar &  & \gb{lex}, \gb{sum} \\
\rowcolor{vlgray}{} Number Partitioning & & \gb{allDifferent}, \gb{sum} \\
Rostering &  & \gb{allDifferent}, \gb{regular} \\
\rowcolor{vlgray}{} Ortho. Latin Squares & & \gb{allDifferent}, \gb{table} \\
Pseudo-Boolean &  & \gb{sum} \\
\rowcolor{vlgray}{} Quasigroup & & \gb{allDifferent}, \gb{element} \\
Room Mate &  &  \\
\rowcolor{vlgray}{} Solitaire Battleship & & \gb{cardinality}, \gb{count}, \gb{regular}, \gb{sum}, \gb{table} \\
Sports Scheduling & & \gb{allDifferent}, \gb{cardinality}, \gb{count}, \gb{table} \\
\rowcolor{vlgray}{} Superpermutation & & \gb{allDifferent}, \gb{element} \\
& & \\
\midrule
COP Problems & Optimization & Global Constraints\\
\midrule
\rowcolor{vlgray}{} Aircraft Landing &  $\min$ EXPR & \gb{allDifferent}, \gb{noOverlap}, \gb{table}  \\
 Clock Triplets &  $\min$ VAR & \gb{allDifferent}, \gb{sum} \\
\rowcolor{vlgray}{} Coins Grid &  $\min$ SUM & \gb{sum} \\
CVRP & $\min$ SUM & \gb{allDifferent}, \gb{cardinality}, \gb{element}, \gb{sum} \\
\rowcolor{vlgray}{} Cyclic Bandwith &  $\min$ MAXIMUM & \gb{allDifferent} \\
DC & $\min$ SUM & \gb{extension}, \gb{sum} \\
\rowcolor{vlgray}{} Echelon Stock & $\min$ SUM & \gb{sum} \\
Filters & $\min$ MAXIMUM & \gb{noOverlap} \\
\rowcolor{vlgray}{} Itemset Mining & $\max$ SUM & \gb{count}, \gb{lex}, \gb{sum} \\
Multi-Agent Path Finding &  $\min$ MAXIMUM & \gb{allDifferent}, \gb{table} \\
& $\min$ SUM & \\
\rowcolor{vlgray}{} Nurse Rostering & $\min$ SUM & \gb{count}, \gb{regular}, \gb{slide}, \gb{sum}, \gb{table} \\
Nursing Workload & $\min$ SUM & \gb{cardinality}, \gb{sum} \\
\rowcolor{vlgray}{} RCPSP & $\min$ VAR & \gb{cumulative} \\
RLFAP & $\min$ MAXIMUM &  \\
& $\min$ N\_VALUES & \\
& $\min$ SUM & \\
\rowcolor{vlgray}{} Spot5 &  $\min$ SUM & \gb{table} \\
TAL & $\min$ SUM & \gb{count}, \gb{table} \\
\rowcolor{vlgray}{} Triangular & $\max$ SUM & \gb{sum} \\
Warehouse & $\min$ SUM & \gb{count}, \gb{element} \\
\rowcolor{vlgray}{} War or Peace & $\min$ SUM & \gb{sum} \\
\bottomrule
  \end{tabular}
  \end{small}
  \caption{Selected Problems for the main tracks of the 2022 Competition. VAR/EXPR means that a variable/expression must be optimized. For RLFAP and Multi-agent Path Finding, the type of objective differs depending on instances. When \gb{extension} is followed by ($*$), it means that short tables are involved.}\label{fig:problems}
\end{table}

\paragraph{Novelty.} Many problems are new in 2022, with most of the models written in \p3. Two problems have been submitted, in response to the call. 

\paragraph{Diversity.} Of course, not all types of constraints are equally involved in the selected benchmark.
In the next edition, we shall attempt to foster the representation of CP-representative global constraints such as \gb{cumulative} and \gb{noOverlap}, and possibly, to introduce a couple of new ones (e.g., \gb{binPacking}).

\paragraph{Scaling up.} It is always interesting to see how constraint solvers behave when the instances of a problem become harder and harder.
This is what we call the scaling behavior of solvers.
For most of the problems in the 2022 competition, we have selected series of instances with regular increasing difficulty.
It is important to note that assessing the difficulty of instances was determined with \ace, which is the main reason why \ace is off-competition (due to this strong bias).

\paragraph{Selection.} This year, the selection of problems and instances has been performed by Christophe Lecoutre.
As a consequence, the solver \ace was labelled off-competition.


\bigskip

\chapter{Problems and Models}

In the next sections, you will find all models used for generating the \x3 instances of the 2022 competition (for main CSP and COP tracks).
Almost all models are written in \p3 \cite{LS_pycsp3}, Version 2.0, officially released in December 2021; see \href{https://pycsp.org/}{https://pycsp.org/}.

\section{CSP}

\subsection{Aztec Diamond}

\paragraph{Description.}
An Aztec diamond of order $n$ consists of $2n$ centered rows of unit squares, of respective lengths $2, 4, \dots, 2n-2, 2n, 2n-2, \dots, 4, 2$.
An Aztec diamond of order $n$ has exactly $2^{(n*(n+1)/2)}$ tilings by dominos.
See \href{https://en.wikipedia.org/wiki/Aztec_diamond}{wikipedia.org}.
Building a solution analytically may be easy. 
However, a CP model is interesting as one can easily add side constraints to form Aztec diamonds with some specific properties (although this is not the case here).

\paragraph{Data.}

Only one integer is required to specify a specific instance. Values of $n$ used for the competition are:
\begin{quote}
25, 50, 75, 100, 150, 200, 250, 300
\end{quote}

\paragraph{Model.}
The \p3 model, in a file `AztecDiamond.py', used for the competition is:

\begin{boxpy}\begin{python}
@\imp@

n = data  # order of the Aztec diamond

def valid(i, j):
   if i < 0 or i >= n * 2 or j < 0 or j >= n * 2:
      return False
   if i < n - 1 and (j < n - 1 - i or j > n + i):
      return False
   if i > n and (j < i - n or j > 3 * n - i - 1):
      return False
   return True

def inner(i, j):
    return valid(i, j - 1) and valid(i, j + 1) and valid(i - 1, j) and valid(i + 1, j)
    
# all valid cells
valid_cells = [(i, j) for i in range(2 * n) for j in range(2 * n) if valid(i, j)]

# all inner cells, i.e., valid cells that are not situated on the border of the diamond
inners = [(i, j) for i, j in valid_cells if inner(i, j)]

# all border cells, i.e., valid cells that are situated on the border of the diamond
borders = [(i, j) for i, j in valid_cells if not inner(i, j)]

T =  {(0, 1, ANY, ANY, ANY), (1, ANY, 0, ANY, ANY), (2, ANY, ANY, 3, ANY), (3, ANY,ANY,ANY,2)}

# x[i][j] is the position (0: left, 1: right, 2:top, 3: bottom) of the second part of the domino whose first part occupies the cell ar row i and column j
x = VarArray(size=[2 * n, 2 * n], dom=lambda i, j: range(4) if valid(i, j) else None)

satisfy(
   # constraining cells situated on the top left border
   [(x[i][j], x[i][j + 1], x[i + 1][j]) in {(1, 0, ANY), (3, ANY, 2)}
      for i, j in borders if not valid(i, j - 1) and not valid(i - 1, j)],

   # constraining cells situated on the top right border
   [(x[i][j], x[i][j - 1], x[i + 1][j]) in {(0, 1, ANY), (3, ANY, 2)}
      for i, j in borders if not valid(i, j + 1) and not valid(i - 1, j)],
  
   # constraining cells situated on the bottom left border
   [(x[i][j], x[i][j + 1], x[i - 1][j]) in {(1, 0, ANY), (2, ANY, 3)}
      for i, j in borders if not valid(i, j - 1) and not valid(i + 1, j)],
  
   # constraining cells situated on the bottom right border
   [(x[i][j], x[i][j - 1], x[i - 1][j]) in {(0, 1, ANY), (2, ANY, 3)}
      for i, j in borders if not valid(i, j + 1) and not valid(i + 1, j)],
  
   # constraining inner cells
   [(x[i][j], x[i][j - 1], x[i, j + 1], x[i - 1][j], x[i + 1][j]) in T
      for i, j in inners]
)
\end{python}\end{boxpy}

The model involves a two-dimensional array of variable $x$, and several groups (lists) of starred table constraints (ANY is '*').
A series of 8 instances has been selected for the competition.
For generating an \x3 instance (file), you can execute for example:
\begin{command}
python AztecDiamond.py -data=100
\end{command}

\subsection{Blocked Queens}

This is Problem \href{https://www.csplib.org/Problems/prob080/}{080} on CSPLib, called Blocked n-Queens Problem.

\paragraph{Description {\small (excerpt from CSPLib)}.}
The blocked n-queens problem is a variant of n-queens which has been proven to be NP-complete as a decision problem and \#P-complete as a counting problem.
The blocked n-queens problem is a variant where, as well as n, the input contains a list of squares which are blocked. A solution to the problem is a solution to the n-Queens problem containing no queens on any of the blocked squares.

\paragraph{Data.}
As an illustration of data specifying an instance of this problem, we have:
\begin{json}
{
  "n": 6,
  "blocks": [[0,2], [3,4], [5,1]]
}
\end{json}

A series of 400 instances (not in JSON format) can be found on CSPLib.
They were used in ASP competitions.
The problem was also the subject of the LP/CP programming contest at ICLP 2016. 

\paragraph{Model.}
The \p3 model, in a file `BlockedQueens.py', used for the competition is:

\begin{boxpy}\begin{python}
@\imp@

n, blocks = data

# q[i] is the column where is put the ith queen (at row i)
q = VarArray(size=n, dom=range(n))

satisfy(
   # respecting blocks
   [q[i] != j for (i, j) in blocks],

   # no two queens on the same column
   AllDifferent(q),

   # no two queens on the same diagonal
   [abs(q[i] - q[j]) != abs(i - j) for i, j in combinations(n, 2)]
)
\end{python}\end{boxpy}

The model involves a global constraint \gb{AllDifferent} and some unary and binary intensional constraints.
A series of 8 instances has been selected for the competition.
For generating an \x3 instance (file), you can execute for example:
\begin{command}
python BlockedQueens.py -data=28-1449787798 -parser=BlockedQueens_Parser.py
\end{command}

where `28-1449787798' is a data file in Essence format and `BlockedQueens\_Parser.py' is a parser (i.e., a Python file allowing us to load data that are not directly given in JSON format).
Note that for saving data in JSON files, you can add the option `-export' (or `-dataexport').

\subsection{Car Sequencing}

This is Problem \href{https://www.csplib.org/Problems/prob001/}{001} on CSPLib. 

\paragraph{Description {\small (excerpt from CSPLib)}.}
A number of cars are to be produced; they are not identical, because different options are available as variants on the basic model. The assembly line has different stations which install the various options (air-conditioning, sun-roof, etc.). These stations have been designed to handle at most a certain percentage of the cars passing along the assembly line. Furthermore, the cars requiring a certain option must not be bunched together, otherwise the station will not be able to cope. Consequently, the cars must be arranged in a sequence so that the capacity of each station is never exceeded. For instance, if a particular station can only cope with at most half of the cars passing along the line, the sequence must be built so that at most 1 car in any 2 requires that option.

\paragraph{Data.}

As an illustration of data specifying an instance of this problem, we have:
\begin{json}
{
  "carClasses": [
    { "demand": 1, "options": [1, 0, 1, 1, 0] },
    { "demand": 2, "options": [0, 1, 0, 0, 1] },
    { "demand": 2, "options": [0, 1, 0, 1, 0] },
    { "demand": 2, "options": [1, 1, 0, 0, 0] }
  ],
  "optionLimits": [
    { "num": 1, "den": 2 },
    { "num": 2, "den": 3 },
    { "num": 1, "den": 3 },
    { "num": 2, "den": 5 },
    { "num": 1, "den": 5 }
  ]
}
\end{json}

\paragraph{Model.}

The \p3 model, in a file `CarSequencing.py', used for the competition is:

\begin{boxpy}\begin{python}
@\imp@

classes, limits = data
demands, options = zip(*classes) 
nCars, nClasses, nOptions = sum(demands), len(classes), len(limits)
   
# c[i] is the class of the ith assembled car
c = VarArray(size=nCars, dom=range(nClasses))

# o[i][k] is 1 if the ith assembled car has option k
o = VarArray(size=[nCars, nOptions], dom={0, 1})

def sum_from_full_consecutive_blocks(k, nb):
   # nb stands for the number of consecutive blocks (of cars) set to their maximal capacity
   n_cars_with_option = sum(demand for (demand, opts) in classes if opts[k] == 1)
   remaining = n_cars_with_option - nb * limits[k].num
   possible = nCars - nb * limits[k].den
   return Sum(o[:possible, k]) >= remaining if remaining > 0 and possible > 0 else None 

satisfy(
   # building the right numbers of cars per class
   Cardinality(c, occurrences=demands) 
)

if not variant():
   satisfy(
      # computing assembled car options
      If(
         c[i] == j,
         Then=o[i] == options[j]
      ) for i in range(nCars) for j in range(nClasses)
   )
elif variant('table'):
   satisfy(
      # computing assembled car options
      (c[i], o[i]) in enumerate(options) for i in range(nCars)
   )

satisfy(
   # respecting option frequencies
   [Sum(o[i:i + den, k]) <= num for k, (num, den) in enumerate(limits)
      for i in range(nCars) if i <= nCars - den],

   
   # additional constraints by reasoning from consecutive blocks  tag(redundant-constraints)
   [sum_from_full_consecutive_blocks(k, nb) for k in range(nOptions)
      for nb in range(ceil(nCars // limits[k].den) + 1)]
)
\end{python}\end{boxpy} 

This model involves 2 arrays of variables and 4 types of constraints: \gb{Cardinality}, \gb{Intension}, \gb{Extension} and \gb{Sum}.
Actually, depending on the chosen variant, either \gb{Extension} constraints are posted, or binary \gb{Intension} constraints are posted with predicates like $c_i = j \Rightarrow o_{i,k} = v$ where $v$ is the value (0 or 1) of the kth option of the jth class.
The last group of constraints corresponds to redundant constraints.
A series of 10 instances has been selected for the competition (8 for variant `table').
For generating an \x3 instance (file), you can execute for example:
\begin{command}
python CarSequencing.py -data=90-01 -parser=CarSequencing_Parser.py
python CarSequencing.py -data=90-01 -parser=CarSequencing_Parser.py -variant=table
\end{command}

where `90-01' is a data file and `CarSequencing\_Parser.py' is a parser (i.e., a Python file allowing us to load data that are not directly given in JSON format).
Note that when you omit to write `-variant=table', you get the main variant.
Note that for saving data in JSON files, you can add the option `-export'.

\subsection{Costas Arrays}

This is Problem \href{https://www.csplib.org/Problems/prob076/}{076} on CSPLib. 

\paragraph{Description {\small (excerpt from CSPLib)}.}
A costas array is a pattern of $n$ marks on an $n \times n$ grid, one mark per row and one per column, in which the $n \times (n - 1)/2$ vectors between the marks are all different.
Such patterns are important as they provide a template for generating radar and sonar signals with ideal ambiguity functions.

\paragraph{Data.}
Only one integer is required to specify a specific instance. Values of $n$ used for the instances in the competition are:
\begin{quote}
15, 16, 17, 18, 19, 20, 22, 24, 26, 28
\end{quote}

\paragraph{Model.}
The \p3 model, in a file `CostasArray.py', used for the competition is: 

\begin{boxpy}\begin{python}
@\imp@

n = data

# x[i] is the row where is put the ith mark (on the ith column)
x = VarArray(size=n, dom=range(n))

satisfy(
   # all marks are on different rows (and columns)
   AllDifferent(x),

   # all displacement vectors between the marks must be different
   [AllDifferent(x[i] - x[i + d] for i in range(n - d)) for d in range(1, n - 1)]
)
\end{python}\end{boxpy} 

This model involves 1 array of variables and several constraints \gb{AllDifferent}.
A series of 10 instances has been selected for the competition.
For generating an \x3 instance (file), you can execute for example:
\begin{command}
python CostasArray.py -data=20
\end{command}

\subsection{Crosswords (Satisfaction)}

This problem has already been used in previous XCSP competitions, because it notably permits to compare filtering algorithms on large table constraints.

\paragraph{Description.}
Given a grid with imposed black cells (spots) and a dictionary, the problem is to fulfill the grid with the words contained in the dictionary.

\paragraph{Data.}
As an illustration of data specifying an instance of this problem, we have:
\begin{json}
{
  "spots": [[0,1,0,0,0], [0,0,0,0,0], [0,0,1,0,0], [0,0,0,0,0], [0,0,0,0,1]],
  "dict": "ogd2008"
}
\end{json}

\paragraph{Model.}
A model similar to the following one, in a file `Crossword.py', was used in 2018: 

\begin{boxpy}\begin{python}
@\imp@

spots, dict_name = data

# loading words of the dictionary
words = dict()
for line in open(dict_name):
   code = alphabet_positions(line.strip().lower())
   words.setdefault(len(code), []).append(code)

# For Hole, i and j are indexes (one of them being a slice) and r is the size
Hole = namedtuple("Hole", "i j r")

def build_hole(row, col, size, horizontal):
   sl = slice(col, col + size)
   return Hole(row, sl, size) if horizontal else Hole(sl, row, size)

def find_holes(matrix, transposed):
   p, q = len(matrix), len(matrix[0])
   t = []
   for i in range(p):
      start = -1
      for j in range(q):
         if matrix[i][j] == 1:
            if start != -1 and j - start >= 2:
               t.append(build_hole(i, start, j - start, not transposed))
            start = -1
         elif start == -1:
            start = j
         elif j == q - 1 and q - start >= 2:
            t.append(build_hole(i, start, q - start, not transposed))
   return t

holes = find_holes(spots, False) + find_holes(columns(spots), True)
n, m, nHoles = len(spots), len(spots[0]), len(holes)

#  x[i][j] is the letter, number from 0 to 25, at row i and column j (when no spot)
x = VarArray(size=[n, m], dom=lambda i, j: range(26) if spots[i][j] == 0 else None)

satisfy(
   # fill the grid with words
   x[i, j] in words[r] for (i, j, r) in holes
)    
\end{python}\end{boxpy}

This model only involves 1 array of variables and 1 group of ordinary table constraints.
For clarity, we use an auxiliary named tuple \texttt{Hole}.
A series of 13 instances, with only blank grids, has been selected for the competition.
Note that it is not possible to write x[i][j] when i is a slice; this must be x[i, j].
For generating an \x3 instance (file), you can execute for example:
\begin{command}
python Crossword.py -data=[vg0405,dict=ogd2008] -parser=Crossword_Parser.py 
\end{command}
where `vg0405' is a data file representing a grid, `ogd2008' is the name of a dictionary file (we need to use 'dict=' as a prefix, otherwise the dictionary will be appended to the first file) and `Crossword\_Parser.py' is a parser (i.e., a Python file allowing us to load data that are not directly given in JSON format).
Note that for saving data in JSON files, you can add the option `-export'.

{\em Important:} The series of instances, used for the 2022 competition comes from the 2018 competition, and compiling instances from the \p3 model above may produce slighlty different files.

\subsection{Crypto}

A series of 10 instances has been generated  (independently of \p3), and submitted to the 2022 competition by Martin Mariusz Lester.
This benchmark encodes 10 instances of breaking the weak stream cipher \href{https://en.wikipedia.org/wiki/Crypto-1}{Crypto1}.
M. M. Lester generated the instances using the tool \href{https://www.msoos.org/grain-of-salt/}{Grain of Salt} (which produces CNF format SAT instances), then translated them into \x3 instances.
Martin picked instances that were relatively harder for MiniSAT to solve (2-4 minutes).
He hypothesised that these instances would be hard for any \x3 solver that does not use a SAT solver as the backend, with \x3 solvers using more modern solvers faring best.
Perhaps this benchmark is not very interesting from the perspective of using \x3 as an intermediate language, as it is a translation back from a low-level language that loses all of the structure.
Nonetheless, it may serve as a reminder that these kinds of encodings exist.

\subsection{Diamond Free}

This is Problem \href{https://www.csplib.org/Problems/prob050/}{050} on CSPLib, called Diamond-free Degree Sequences. 

\paragraph{Description {\small (excerpt from CSPLib)}.}
A diamond is a set of four vertices in a graph such that there are at least five edges between those vertices.
Conversely, a graph is diamond-free if it has no diamond as an induced subgraph, i.e. for every set of four vertices the number of edges between those vertices is at most four.

\paragraph{Data.}
Only one integer is required to specify a specific instance. Values of $n$ used for the instances in the competition are:
\begin{quote}
30, 40, 50, 60, 70, 80
\end{quote}

\paragraph{Model.}
The \p3 model, in a file `DiamondFree.py', used for the competition is:

\begin{boxpy}\begin{python}
@\imp@

n = data

# x is the adjacency matrix
x = VarArray(size=[n, n], dom=lambda i, j: {0, 1} if i != j else {0})

# y[i] is the degree of the ith node
y = VarArray(size=n, dom={i for i in range(1, n) if i % 3 == 0})

# s is the sum of all degrees
s = Var(dom={i for i in range(n, n * (n - 1) + 1) if i % 12 == 0})

satisfy(
   # ensuring the absence of diamond in the graph
   [Sum(x[i][j], x[i][k], x[i][l], x[j][k], x[j][l], x[k][l]) <= 4
      for i, j, k, l in combinations(n, 4)],

   # ensuring that the graph is undirected (symmetric)
   [x[i][j] == x[j][i] for i, j in combinations(n, 2)],
   
   # computing node degrees
   [Sum(x[i]) == y[i] for i in range(n)],
   
   # computing the sum of node degrees
   Sum(y) == s,
   
   # tag(symmetry-breaking)
   [
      Decreasing(y),
      LexIncreasing(x)
   ]
)
\end{python}\end{boxpy}

This model involves 2 arrays of variables, a stand-alone variable and 4 types of constraints: \gb{Sum}, \gb{Intension}, \gb{Decreasing} and \gb{LexIncreasing}.
A series of 6 instances has been selected for the competition.
For generating an \x3 instance (file), you can execute for example:
\begin{command}
python DiamondFree.py -data=40
\end{command}

\subsection{Eternity}

Eternity II is a famous edge-matching puzzle, released in July 2007 by TOMY, with a 2 million dollars prize for the first submitted solution; see, e.g., \cite{BB_fast}.
Here, we are interested in instances derived from the original problem by the \href{http://becool.info.ucl.ac.be/}{BeCool} team of the UCL (``Universit\'e Catholique de Louvain'') who proposed them for the 2018 competition.

\paragraph{Description.}
On a board of size $n \times m$, you have to put square tiles (pieces) that are described by four colors (one for each direction : top, right, bottom and left).
All adjacent tiles on the board must have matching colors along their common edge. All edges must have color '0' on the border of the board.

\paragraph{Data.}
As an illustration of data specifying an instance of this problem, we have:
\begin{json}
{
  "n": 3,
  "m": 3,
  "pieces": [
     [0,0,1,1], [0,0,1,2], [0,0,2,1], [0,0,2,2], [0,1,3,2],
     [0,1,4,1], [0,2,3,1], [0,2,4,2], [3,3,4,4]
  ]
}
\end{json}

\paragraph{Model.}
The \p3 model, in a file `Eternity.py', used for the competition is:

\begin{boxpy}\begin{python}
@\imp@

n, m, pieces = data
assert n * m == len(pieces), "badly formed data"
max_value = max(max(piece) for piece in pieces)  # max possible value on pieces

T = {(i, piece[r % 4], piece[(r + 1) % 4], piece[(r + 2) % 4], piece[(r + 3) % 4])
            for i, piece in enumerate(pieces) for r in range(4)}

# x[i][j] is the index of the piece at row i and column j
x = VarArray(size=[n, m], dom=range(n * m))

# t[i][j] is the value at the top of the piece at row i and column j
t = VarArray(size=[n + 1, m], dom=range(max_value + 1))

# l[i][j] is the value at the left of the piece at row i and column j
l = VarArray(size=[n, m + 1], dom=range(max_value + 1))

satisfy(
   # all pieces must be placed (only once)
   AllDifferent(x),

   # all pieces must be valid (i.e., must correspond to those given initially,
   #  possibly after applying some rotation)
   [(x[i][j], t[i][j], l[i][j + 1], t[i + 1][j], l[i][j]) in T
      for i in range(n) for j in range(m)],
   
   # putting special value 0 on borders
   [z == 0 for z in t[0] + l[:, -1] + t[-1] + l[:, 0]]
)
\end{python}\end{boxpy} 

This model involves 3 arrays of variables and 3 types of constraints: \gb{AllDifferent}, \gb{Extension} and \gb{Intension}.
A series of 10 instances has been selected for the competition.
For generating an \x3 instance (file), you can execute for example:
\begin{command}
python Eternity.py -data=06-06 -parser=Eternity_Parser.py 
\end{command}
where `06-06' is a data file representing a puzzle and `Eternity\_Parser.py' is a parser (i.e., a Python file allowing us to load data that are not directly given in JSON format).
Note that for saving data in JSON files, you can add the option `-export'.

\subsection{Hadamard}

This is Problem \href{https://www.csplib.org/Problems/prob084/}{084} on CSPLib, called 2cc Hadamard matrix Legendre pairs.

\paragraph{Description.}
For every odd positive integer $n$ (and $m = (n-1)/2$), the 2cc Hadamard matrix Legendre pairs are defined from $m$ quadratic constraints and 2 linear constraints.

\paragraph{Data.}
Only one integer is required to specify a specific instance. Values of $n$ used for the instances in the competition are:
\begin{quote}
17, 19, 21, 23, 25, 27, 29, 31, 35, 41
\end{quote}

\paragraph{Model.}
The \p3 model, in a file `Hadamard.py', used for the competition is: 

\begin{boxpy}\begin{python}
@\imp@

n = data 
assert n % 2 == 1
m = (n - 1) // 2

# x[i] is the ith value of the first sequence
x = VarArray(size=n, dom={-1, 1})

# y[i] is the ith value of the second sequence
y = VarArray(size=n, dom={-1, 1})

satisfy(
   Sum(x) == 1,
   Sum(y) == 1,

   # quadratic constraints
   [
      Sum(x[i] * x[i + k] for i in range(n)) + Sum(y[i] * y[i + k] for i in range(n)) == -2
         for k in range(1, m + 1)
   ]
)

\end{python}\end{boxpy}

This model involves 2 arrays of variables and several constraints \gb{Sum}.
Note that, by auto-adjustment of array indexing, \verb!x[i + k]! is equivalent to \verb!x[(i + k) % n]!.
A series of 10 instances has been selected for the competition.
For generating an \x3 instance (file), you can execute for example:
\begin{command}
python Hadamard.py -data=23
\end{command}

\subsection{Hidato}

\paragraph{Description.}
Hidato, also known as Hidoku is a logic puzzle game invented by Gyora M. Benedek, an Israeli mathematician.
The goal of Hidato is to fill the grid with consecutive numbers that connect horizontally, vertically, or diagonally.
See \href{https://en.wikipedia.org/wiki/Hidato}{wikipedia.org}.

\paragraph{Data.}
As an illustration of data specifying an instance of this problem, we have:
\begin{json}
{
  "n": 5,
  "m": 5,
  "clues": [
    [0, 0, 20, 0, 0],
    [0, 0, 0, 16, 18],
    [22, 0, 15, 0, 0],
    [23, 0, 1, 14, 11],
    [0, 25, 0, 0, 12]
  ]
}
\end{json}

\paragraph{Model.}
The \p3 model, in a file `Hidato.py', used for the competition is: 

\begin{boxpy}\begin{python}
@\imp@

n, m, clues = data   # clues are given by strictly positive values

# x[i][j] is the value in the grid at row i and column j
x = VarArray(size=[n, m], dom=range(1, n * m + 1))

satisfy(
   # all values must be different
   AllDifferent(x),

   # respecting clues
   [x[i][j] == clues[i][j] for i in range(n) for j in range(m) if clues and clues[i][j] > 0]
)

if variant('table'):
   
   def table(i, j):
      corners = {(0, 0), (0, m - 1), (n - 1, 0), (n - 1, m - 1)}
      r = 3 if (i, j) in corners else 5 if i in (0, n - 1) or j in (0, m - 1) else 8
      return  [(v, *[v + 1 if l == k else ANY for l in range(r)])
                 for v in range(1, n * m) for k in range(r)]
              + [(n * m, *[ANY] * r)]
   
   satisfy(
      # ensuring adjacent consecutive numbers
      (x[i][j], x.around(i, j)) in table(i, j) for i in range(n) for j in range(m)
   )

else:  # variant not considered for the competition
   ...   
\end{python}\end{boxpy} 

This model involves 1 array of variables and 3 types of constraints: \gb{AllDifferent}, \gb{Extension} and \gb{Intension}.
A series of 9 instances (with variant `table') has been selected for the competition.
For generating an \x3 instance (file), you can execute for example:
\begin{command}
python Hidato.py -variant=table -data=p1.json
\end{command}
or with an empty grid of size $10 \times 10$:
\begin{command}
python Hidato.py -variant=table -data=[10,10,null] 
\end{command}

\subsection{Knight Tour}

\paragraph{Description.}
A knight's tour is a sequence of moves of a knight on a chessboard such that the knight visits every square exactly once.
If the knight ends on a square that is one knight's move from the beginning square (so that it could tour the board again immediately, following the same path), the tour is closed (or re-entrant); otherwise, it is open.
See \href{https://en.wikipedia.org/wiki/Knight%27s_tour}{wikipedia.org}.

\paragraph{Data.}
Only one integer is required to specify a specific instance. Values of $n$ used for the instances in the competition are:
\begin{quote}
10, 20, 30, 40, 50, 60, 70, 80, 90, 100
\end{quote}

\paragraph{Model.}
The \p3 model, in a file `KnightTour2.py', used for the competition is: 

\begin{boxpy}\begin{python}
@\imp@

n = data 

def domain_x(i):
   r, c = i // n, i % n
   t = [(r - 2, c - 1), (r - 2, c + 1), (r - 1, c - 2), (r - 1, c + 2),
        (r + 1, c - 2), (r + 1, c + 2), (r + 2, c - 1), (r + 2, c + 1)]
   return {k * n + l for (k, l) in t if 0 <= k < n and 0 <= l < n}

# x[i] is the cell number that comes in the tour (by the knight) after cell i
x = VarArray(size=n * n, dom=domain_x)

satisfy(
   # the knights form a circuit (tour)
   Circuit(x),

   # the first move is set  tag(symmetry-breaking)
   x[0] == n + 2
)
\end{python}\end{boxpy} 

This model involves 1 array of variables and 2 types of constraints: \gb{Circuit} and \gb{Intension}.
A series of 10 instances has been selected for the competition.
For generating an \x3 instance (file), you can execute for example:
\begin{command}
python KnightTour2.py -data=50
\end{command}

\subsection{Molnar}

This is Problem \href{https://www.csplib.org/Problems/prob035/}{035} on CSPLib.

\paragraph{Description {\small (excerpt from CSPLib)}.}
The problem (in this variant) is to construct a $k \times k$ matrix $M$ with values strictly greater than 1
such that both the determinant of $M$ and the determinant of $M$ when squared values are considered are equal to 1  
The solutions to this problem are significant in classifying certain types of topological spaces. 

\paragraph{Data.}
Only two integers are required to specify a specific instance. Values of $k$ and $d$ used for the instances in the competition are:
\begin{quote}
(2,20), (2,25), (3,7), (3,8), (3,9), (4,3), (4,4), (4,5), (5,4), (5,5)
\end{quote}

\paragraph{Model.}
The \p3 model, in a file `Molnar.py', used for the competition is: 

\begin{boxpy}\begin{python}
@\imp@
@\textcolor{mgray}{from pycsp3.classes.entities import TypeNode}@

k, d = data

def det_terms(t):
   if len(t) == 2:
      return [t[0][0] * t[1][1], -(t[0][1] * t[1][0])]
   subterms = [det_terms([[v for j, v in enumerate(row) if j != i] for row in t[1:]])
                   for i in range(len(t))]
   return [t[0][i] * sub if i % 2 == 0 else -(t[0][i] * sub) for i in range(len(t))
              for sub in subterms[i]]

def determinant(t):
   terms = det_terms(t)
   # we extract coeffs from terms for posting a simpler Sum constraint later
   terms = [(term.sons[0], -1) if term.type == TypeNode.NEG else (term, 1) for term in terms]
   return [t for t, _ in terms] * [c for _, c in terms]


# x[i][j] is the value of the matrix at row i and column j
x = VarArray(size=[k, k], dom=range(2, d + 1))

# y[i][j] is the square of the value of the matrix x at row i and column j
y = VarArray(size=[k, k], dom=range(4, d * d + 1))

satisfy(
   # computing y
   [y[i][j] == x[i][j] * x[i][j] for i in range(k) for j in range(k)],

   determinant(x) == 1,

   determinant(y) == 1,

   # tag(symmetryBreaking)
   LexIncreasing(x, matrix=True)
)
\end{python}\end{boxpy} 

This model involves 2 arrays of variables and 3 types of constraints: \gb{Sum}, \gb{LexIncreasing} and \gb{Intension}.
A series of 10 instances has been selected for the competition.
For generating an \x3 instance (file), you can execute for example:
\begin{command}
python Molnar.py -data=[4,5]
\end{command}

\subsection{Number Partitioning}

This is Problem \href{https://www.csplib.org/Problems/prob049/}{049} on CSPLib.

\paragraph{Description {\small (excerpt from CSPLib)}.}
This problem consists in finding a partition of the set of numbers $\{1,2,\dots,n\}$ into two sets A and B such that:
\begin{itemize}
\item A and B have the same cardinality
\item the sum of numbers in A is equal to the sum of numbers in B
\item the sum of squares of numbers in A is equal to the sum of squares of numbers in B
\end{itemize}
There is no solution for $n<8$.

\paragraph{Data.}
Only one integer is required to specify a specific instance. Values of $n$ used for the instances in the competition are:
\begin{quote}
20, 50, 80, 110, 140, 170, 200, 230, 260, 290
\end{quote}

\paragraph{Model.}
The \p3 model, in a file `NumberPartitioning.py', used for the competition is: 

\begin{boxpy}\begin{python}
@\imp@

n = data 
assert n % 2 == 0, "The value of n must be even"
K1, K2 = n * (n + 1) // 4, n * (n + 1) * (2 * n + 1) // 12

# x[i] is the ith value of the first set
x = VarArray(size=n // 2, dom=range(1, n + 1))

# y[i] is the ith value of the second set
y = VarArray(size=n // 2, dom=range(1, n + 1))

satisfy(
   AllDifferent(x + y),

   # tag(power1)
   [
      Sum(x) == K1,
      Sum(y) == K1
   ],

   # tag(power2)
   [
      Sum(x[i] * x[i] for i in range(n // 2)) == K2,
      Sum(y[i] * y[i] for i in range(n // 2)) == K2
   ],

   # tag(symmetry-breaking)
   [
      x[0] == 1,
      Increasing(x, strict=True),
      Increasing(y, strict=True)
   ]
)
\end{python}\end{boxpy} 

This model involves 2 arrays of variables and 3 types of constraints: \gb{AllDifferent}, \gb{Sum} and \gb{Increasing}.
A series of 10 instances has been selected for the competition.
For generating an \x3 instance (file), you can execute for example:
\begin{command}
python NumberPartitioning.py -data=50
\end{command}

\subsection{(Nurse) Rostering}

This problem was described by G. Pesant, C.-G. Quimper and A. Zanarini \cite{PQZ_counting}

\paragraph{Description.}
This problem was inspired by a rostering context.
The objective is to schedule $n$ employees over a span of $n$ time periods.
In each time period, $n - 1$ tasks need to be accomplished and one employee out of the $n$ has a break.
The tasks are fully ordered 1 to $n - 1$; for each employee the schedule has to respect the following rules:
\begin{itemize}
\item two consecutive time periods have to be assigned to either two consecutive tasks, in no matter which order i.e. (t, t + 1) or (t + 1, t), or to the same task i.e. (t, t);
\item an employee can have a break after no matter which task;
\item after a break an employee cannot perform the task that precedes the task prior to the break, i.e. (t, break, t - 1) is not allowed.
\end{itemize}
The problem is modeled with one constraint \gb{Regular} per row and one constraint \gb{Alldifferent} per column.

\paragraph{Data.}
As an illustration of data specifying an instance of this problem, we have:
\begin{json}
{
  "preset": [[7, 5, 8], [7, 0, 8], [3, 5, 5], [0, 5, 4], [1, 7, 7]],
  "forbidden": []
}
\end{json}

\paragraph{Model.}

The \p3 model, in a file `Rostering.py', used for the competition is: 

\begin{boxpy}\begin{python}
@\imp@

n = 10
preset, forbidden = data

def automaton():
   # q(1,i) means before break and just after reading i
   # q(2,i) means just after reading break (0) and i before
   # q(3,i) means after break and just after reading i
   q, rng = Automaton.q, range(1, n)
   t = [(q(0), 0, q(2, 0))] + [(q(2, 0), i, q(3, i)) for i in rng]
   t.extend((q(0), i, q(1, i)) for i in rng)
   # BE CAREFUL: rule made stricter below than Pesant's rule
   t.extend((q(1, i), j, q(1, j)) for i in rng for j in (i - 1, i + 1) if 1 <= j < n)
   t.extend((q(1, i), 0, q(2, i)) for i in rng)
   # BE CAREFUL: rule made stricter below than Pesant's rule
   t.extend((q(2, i), j, q(3, j)) for i in rng for j in rng if abs(i - j) != 1)
   # BE CAREFUL: rule made stricter below than Pesant's rule
   t.extend((q(3, i), j, q(3, j)) for i in rng for j in (i - 1, i + 1) if 1 <= j < n)
   return Automaton(start=q(0), final=[q(2, i) for i in rng] + [q(3, i) for i in rng],
                      transitions=t)

A = automaton()

# x[i][j] is the task (or break) performed by the ith employee at time j
x = VarArray(size=[n, n], dom=range(n))

satisfy(
   [x[i][j] == k for (i, j, k) in preset],

   [x[i][j] != k for (i, j, k) in forbidden],

   [x[i] in A for i in range(n)],

   [AllDifferent(x[:, j]) for j in range(n)]
)
\end{python}\end{boxpy} 


This model involves 1 array of variables and 3 types of constraints: \gb{Regular}, \gb{AllDifferent} and \gb{Intension}.
A series of 10 instances has been selected for the competition.
For generating an \x3 instance (file), you can execute for example:
\begin{command}
python Rostering.py -data=roster-5-00-02.dat -parser=Rostering_Parser.py
\end{command}
where `roster-5-00-02.dat' is a data file and `Rostering\_Parser.py' is a parser (i.e., a Python file allowing us to load data that are not directly given in JSON format).
Note that for saving data in JSON files, you can add the option `-export' (or `-dataexport').

\subsection{Orthogonal Latin Squares}

\paragraph{Description.}
A Latin square of order $n$ is an $n$ by $n$ array filled with $n$ different symbols (for example, values between 1 and $n$),
each occurring exactly once in each row and exactly once in each column.
Two latin squares of the same order $n$ are orthogonal if each pair of elements in the same position occurs exactly once.
The most easy way to see this is by concatenating elements in the same position and verify that no pair appears twice.
There are orthogonal latin squares of any size except 1, 2, and 6.
See \href{https://en.wikipedia.org/wiki/Mutually_orthogonal_Latin_squares}{wikipedia.org}.

\paragraph{Data.}
Only one integer is required to specify a specific instance. Values of $n$ used for the instances in the competition are:
\begin{quote}
5, 6, 7, 8, 9, 10, 11, 12, 15, 20
\end{quote}

\paragraph{Model.}
The \p3 model, in a file `Ortholatin.py', used for the competition is: 

\begin{boxpy}\begin{python}
@\imp@

n = data

# x is the first Latin square
x = VarArray(size=[n, n], dom=range(n))

# y is the second Latin square
y = VarArray(size=[n, n], dom=range(n))

# z is the matrix used to control orthogonality
z = VarArray(size=[n * n], dom=range(n * n))

table = {(i, j, i * n + j) for i in range(n) for j in range(n)}

satisfy(
   # ensuring that x is a Latin square
   AllDifferent(x, matrix=True),

   # ensuring that y is a Latin square
   AllDifferent(y, matrix=True),
   
   # ensuring that values on diagonals are different  tag(diagonals)
   [AllDifferent(t) for t in [diagonal_down(x), diagonal_up(x),
                                 diagonal_down(y), diagonal_up(y)]],
   
   # ensuring orthogonality of x and y through z
   AllDifferent(z),

   
   # computing z from x and y
   [(x[i][j], y[i][j], z[i * n + j]) in table for i in range(n) for j in range(n)],
   
   # tag(symmetry-breaking)
   [(x[0][j] == j, y[0][j] == j) for j in range(n)]
)
\end{python}\end{boxpy} 

This model involves 3 arrays of variables and 3 types of constraints: \gb{AllDifferent}, \gb{Extension} and \gb{Intension}.
A series of 10 instances has been selected for the competition.
For generating an \x3 instance (file), you can execute for example:
\begin{command}
python Ortholatin.py -data=10
\end{command}

\subsection{PB (Pseudo-Boolean)}

This problem has been already used in previous XCSP competitions.
  
\paragraph{Description.}
Pseudo-Boolean problems generalize SAT problems by allowing linear constraints and, possibly, a linear objective function.

\paragraph{Data.}
As an illustration of data specifying an instance of this problem, we have:

\begin{json}
{
  "n": 144,
  "e": 704,
  "constraints": [
    { "coeffs": [1,1,1,1,1], "nums": [1,17,33,49,81], "op": "=", "limit": 1 },
    { "coeffs": [1,1,1,1,1], "nums": [2,18,34,50,82], "op": "=", "limit": 1 },
    ...
  ],
  "objective": null
}
\end{json}

\paragraph{Model.}
The \p3 model, in a file `PseudoBoolean.py', used for the competition is: 

\begin{boxpy}\begin{python}
@\imp@

n, e, constraints, objective = data  

x = VarArray(size=n, dom={0, 1})

satisfy(
   # respecting each linear constraint
   Sum(x[nums] * coeffs, condition=(op, limit)) for (coeffs, nums, op, limit) in constraints
)
\end{python}\end{boxpy}

This problem involves 1 array of variables and 1 type of constraints: \gb{Sum}.
A series of 11 instances has been selected for the competition.
For generating an \x3 instance (file), you can execute for example:
\begin{command}
python PseudoBoolean.py -data=BeauxArts-K65.opb -parser=PseudoBoolean_Parser.py
\end{command}
where `BeauxArts-K65.opb' is a data file and `PseudoBoolean\_Parser.py' is a parser (i.e., a Python file allowing us to load data that are not directly given in JSON format).
Note that for saving data in JSON files, you can add the option `-export' (or `-dataexport').

\subsection{Quasigroup}

This is Problem \href{https://www.csplib.org/Problems/prob003}{003} on CSPLib, called Quasigroup Existence.

\paragraph{Description {\small (excerpt from CSPLib)}.}
An order $n$ quasigroup is a Latin square of size $n$. That is, a $n \times n$ multiplication table in which each element occurs once in every row and column.
A quasigroup can be specified by a set and a binary multiplication operator, $*$ defined over this set. Quasigroup existence problems determine the existence or non-existence of quasigroups of a given size with additional properties.
For example:
\begin{itemize}
\item QG3: quasigroups for which $(a*b)*(b*a)=a$
\item QG5: quasigroups for which $((b*a)*b)*b=a$
\item QG6: quasigroups for which $(a*b)*b=a*(a*b)$
\end{itemize}
For each of these problems, we may additionally demand that the quasigroup is idempotent. That is, $a*a=a$ for every element $a$.

\paragraph{Data.}
Only one integer is required to specify a specific instance. Values of $n$ used for the instances in the competition are:
\begin{itemize}
\item 8, 9, 10, 11, 16 for variant base-v3
\item 8, 9, 10, 11, 16 for variant base-v5
\item 8, 9, 10, 11, 16 for variant base-v6
\end{itemize}

\paragraph{Model.}
The \p3 model, in a file `QuasiGroup.py', used for the competition is: 

\begin{boxpy}\begin{python}
@\imp@

n = data

# x[i][j] is the value at row i and column j of the quasi-group
x = VarArray(size=[n, n], dom=range(n))

satisfy(
   # ensuring a Latin square
   AllDifferent(x, matrix=True),

   # ensuring idempotence  tag(idempotence)
   [x[i][i] == i for i in range(n)]
)

if variant("base"):
   if subvariant("v3"):
       satisfy(
          x[x[i][j], x[j][i]] == i for i in range(n) for j in range(n)
       )
          
    elif subvariant("v5"):
       satisfy(
          x[x[x[j][i], j], j] == i for i in range(n) for j in range(n)
       )
    elif subvariant("v6"):
       satisfy(
          x[x[i][j], j] == x[i, x[i][j]] for i in range(n) for j in range(n)
       )
elif ...  # not considered for the competition        
\end{python}\end{boxpy}

Three variants of the problem are described here, involving 1 array of variables and various forms of constraints \gb{Element}.
Note the presence of the tag 'idempotence', which easily allows us to activate or deactivate the associated constraints, at parsing time.
A series of $3\times5$ instances has been generated, for problems QG3, QG5 and QG6.

For generating an \x3 instance (file), you can execute for example:
\begin{command}
python QuasiGroup.py -data=10 -variant=base-v3 
python QuasiGroup.py -data=10 -variant=base-v5 
python QuasiGroup.py -data=10 -variant=base-v6 
\end{command}

\subsection{Room Mate}

\paragraph{Description {\small (from Wikipedia)}.}
In mathematics, economics and computer science, the stable-roommate problem is the problem of finding a stable matching for an even-sized set.
A matching is a separation of the set into disjoint pairs (`roommates').
The matching is stable if there are no two elements which are not roommates and which both prefer each other to their roommate under the matching.
This is distinct from the stable-marriage problem in that the stable-roommates problem allows matches between any two elements, not just between classes of "men" and "women". 
See \href{https://en.wikipedia.org/wiki/Stable_roommates_problem}{wikipedia.org}.

\paragraph{Data.}
As an illustration of data specifying an instance of this problem, we have:

\begin{json}
{
  "preferences": [
    [3,4,2,6,5], [6,5,4,1,3], [2,4,5,1,6],
    [5,2,3,6,1], [3,1,2,4,6], [5,1,3,4,2]
  ]
}
\end{json}

\paragraph{Model.}
The \p3 model, in a file `RoomMate.py', used for the competition is: 

\begin{boxpy}\begin{python}
@\imp@

preferences = data
n = len(preferences)

def pref_rank():
   pref = [[0] * n for _ in range(n)]  # pref[i][k] = j <-> guy i has guy j as kth choice
   rank = [[0] * n for _ in range(n)]  # rank[i][j] = k <-> guy i ranks guy j as kth choice
   for i in range(n):
      for k in range(len(preferences[i])):
         j = preferences[i][k] - 1  # because we start at 0
         rank[i][j] = k
         pref[i][k] = j
      rank[i][i] = len(preferences[i])
      pref[i][len(preferences[i])] = i
   return pref, rank

pref, rank = pref_rank()

# x[i] is the value of k, meaning that j = pref[i][k] is the paired agent
x = VarArray(size=n, dom=lambda i: range(len(preferences[i])))

satisfy(
   (
      If(x[i] > rank[i][k], Then=x[k] < rank[k][i]),
      If(x[i] == rank[i][k], Then=x[k] == rank[k][i])
   ) for i in range(n) for k in pref[i] if k != i
)
\end{python}\end{boxpy}

This problem involves 1 array of variables and 1 type of constraints: \gb{Intension}.
A series of 6 instances has been selected for the competition (coming from the \href{http://www.dcs.gla.ac.uk/~pat/roommates/distribution/}{repository} developed by Patrick Prosser).
For generating an \x3 instance (file), you can execute for example:
\begin{command}
python RoomMate.py -data=sr0400.txt -parser=RoomMate_Parser.py
\end{command}
where `sr0400.txt' is a data file and `RoomMate\_Parser.py' is a parser (i.e., a Python file allowing us to load data that are not directly given in JSON format).
Note that for saving data in JSON files, you can add the option `-export' (or `-dataexport').

\subsection{Solitaire Battleship}

This is Problem \href{https://www.csplib.org/Problems/prob014}{014} on CSPLib.

\paragraph{Description {\small (excerpt from CSPLib)}.}
As an illustration, a specific fleet consists of one battleship (four grid squares in length), two cruisers (each three grid squares long), three three destroyers (each two squares long) and four submarines (one square each).
The ships may be oriented horizontally or vertically, and no two ships will occupy adjacent grid squares, not even diagonally.
The digits along the right side of and below the grid indicate the number of grid squares in the corresponding rows and columns that are occupied by vessels.
In each of the puzzles, one or more `shots’ have been taken to start you off. These may show water (indicated by wavy lines), a complete submarine (a circle), or the middle (a square), or the end (a rounded-off square) of a longer vessel.

\paragraph{Data.}
As an illustration of data specifying an instance of this problem, we have:

\begin{json}
{
  "fleet":[{"size":4,"cnt":1}, {"size":3,"cnt":2},
         {"size":2,"cnt":3}, {"size":1,"cnt":4}],
  "hints":[{"type":"c","row":7,"col":10},{"type":"w","row":1,"col":6}],
  "rowSums":[2,4,3,3,2,4,1,1,0,0],
  "colSums":[0,5,0,2,2,3,1,3,2,2]
}
\end{json}

\paragraph{Model.}
The \p3 model, in a file `SolitaireBattleship.py', used for the competition is: 

\begin{boxpy}\begin{python}
@\imp@

fleet, hints, rowSums, colSums = data
surfaces = [ship.size * ship.cnt for ship in fleet]
maxSurf = max(surfaces)
pos, neg = [ship.size for ship in fleet], [-ship.size for ship in fleet]
hints = [] if hints is None else hints
n, nTypes = len(colSums), len(pos)

def automaton(horizontal):
   q = Automaton.q  # for building state names
   t = [(q(0), 0, q(0)), (q(0), neg if horizontal else pos, "qq"), ("qq", 0, q(0))]
   for i in pos:
      v = i if horizontal else -i
      t.append((q(0), v, q(i, 1)))
      t.extend((q(i, j), v, q(i, j + 1)) for j in range(1, i))
      t.append((q(i, i), 0, q(0)))
   return Automaton(start=q(0), final=q(0), transitions=t)

Ah, Av = automaton(True), automaton(False)  

# s[i][j] is 1 iff the cell at row i and col j is occupied by a ship segment
s = VarArray(size=[n + 2, n + 2], dom={0, 1})

# t[i][j] is 0 iff the cell at row i and col j is unoccupied, the type (size) of the ship fragment otherwise; when positive, the ship is put horizontally, vertically otherwise
t = VarArray(size=[n + 2, n + 2], dom=set(neg) | {0} | set(pos))

# cp[i] is the number of positive ship segments of type i
cp = VarArray(size=nTypes, dom=range(maxSurf + 1))

# cn[i] is the number of negative ship segments of type i
cn = VarArray(size=nTypes, dom=lambda i: {0} if fleet[i].size == 1 else range(maxSurf + 1))

def hint_ctr(c, i, j):
   if c == 'w':
      return s[i][j] == 0
   if c in {'c', 'l', 'r', 't', 'b'}:
      return [
         s[i][j] == 1,
         s[i - 1][j] == (1 if c == 'b' else 0),
         s[i + 1][j] == (1 if c == 't' else 0),
         s[i][j - 1] == (1 if c == 'r' else 0),
         s[i][j + 1] == (1 if c == 'l' else 0)
      ]
   if c == 'm':
      return [
         s[i][j] == 1,
         t[i][j] not in {-2, -1, 0, 1, 2},
         (s[i - 1][j], s[i + 1][j], s[i][j - 1], s[i][j + 1]) in {(0, 0, 1, 1), (1, 1, 0, 0)}
      ]

satisfy(
   # no ship on borders
   [(s[0][k] == 0, s[-1][k] == 0, s[k][0] == 0, s[k][-1] == 0) for k in range(n + 2)],

   # respecting the specified row tallies
   [Sum(s[i + 1]) == k for i, k in enumerate(rowSums)],

   
   # respecting the specified column tallies
   [Sum(s[:, j + 1]) == k for j, k in enumerate(colSums)],
   
   # being careful about cells on diagonals
   [(s[i][j], s[i - 1][j - 1], s[i - 1][j + 1], s[i + 1][j - 1], s[i + 1][j + 1])
       in {(0, ANY, ANY, ANY, ANY), (1, 0, 0, 0, 0)}
     for i in range(1, n + 1) for j in range(1, n + 1)],

   # tag(channeling)
   [s[i][j] == (t[i][j] != 0) for i in range(n + 2) for j in range(n + 2)],
   
   # counting the number of occurrences of ship segments of each type
   Cardinality(t[1:n + 1, 1:n + 1], occurrences={pos[i]: cp[i] for i in range(nTypes)}
                                         + {neg[i]: cn[i] for i in range(nTypes)}),
   
   # ensuring the right number of occurrences of ship segments of each type
   [cp[i] + cn[i] == surfaces[i] for i in range(nTypes)],
   
   # ensuring row connectedness of ship segments
   [t[i + 1] in Ah for i in range(n)],
   
   # ensuring column connectedness of ship segments
   [t[:, j + 1] in Av for j in range(n)],
   
   # tag(clues)
   [hint_ctr(c, i, j) for (c, i, j) in hints] 
)
\end{python}\end{boxpy}

This problem involves 4 arrays of variables and 5 types of constraints: \gb{Sum}, \gb{Extension}, \gb{Cardinality}, \gb{Regular} and \gb{Intension}.
A series of 8 instances has been selected for the competition (coming from Minizinc \href{https://github.com/MiniZinc/minizinc-benchmarks/tree/master/solbat}{repository}).
For generating an \x3 instance (file), you can execute for example:
\begin{command}
python SolitaireBattleship.py -data=sb_13_13_5_1.dzn
                              -parser=SolitaireBattleship_ParserZ.py
\end{command}
where `sb\_13\_13\_5\_1.dzn' is a data file and `SolitaireBattleship\_ParserZ.py' is a parser (i.e., a Python file allowing us to load data that are not directly given in JSON format).
Note that for saving data in JSON files, you can add the option `-export' (or `-dataexport').

\subsection{Sports Scheduling}

This is Problem \href{https://www.csplib.org/Problems/prob026}{026} on CSPLib, called the Sports Tournament Scheduling.

\paragraph{Description  {\small (excerpt from CSPLib)}.}
The problem is to schedule a tournament of $n$ teams over $n - 1$ weeks, with each week divided into $n/2$ periods, and each period divided into two slots.
The first team in each slot plays at home, whilst the second plays the first team away.
A tournament must satisfy the following three constraints: every team plays once a week; every team plays at most twice in the same period over the tournament; every team plays every other team.

\paragraph{Data.}
Only one integer is required to specify a specific instance. Values of $n$ used for the instances in the competition are:
\begin{quote}
8, 10, 12, 14, 16, 18, 20, 22, 24, 26
\end{quote}

\paragraph{Model.}
The \p3 model, in a file `SportsScheduling.py', used for the competition is: 

\begin{boxpy}\begin{python}
@\imp@

nTeams = data
nWeeks, nPeriods, nMatches = nTeams - 1, nTeams // 2, (nTeams - 1) * nTeams // 2

def match_number(t1, t2):
   return nMatches - ((nTeams - t1) * (nTeams - t1 - 1)) // 2 + (t2 - t1 - 1)

T = {(t1, t2, match_number(t1, t2)) for t1, t2 in combinations(nTeams, 2)}

# m[w][p] is the number of the match at week w and period p
m = VarArray(size=[nWeeks, nPeriods], dom=range(nMatches))

# x[w][p] is the first team for the match at week w and period p
x = VarArray(size=[nWeeks, nPeriods], dom=range(nTeams))

# y[w][p] is the second team for the match at week w and period p
y = VarArray(size=[nWeeks, nPeriods], dom=range(nTeams))

satisfy(
   # all matches are different (no team can play twice against another team)
   AllDifferent(m),

   # linking variables through ternary table constraints
   [(x[w][p], y[w][p], m[w][p]) in T for w in range(nWeeks) for p in range(nPeriods)],
   
   # each week, all teams are different (each team plays each week)
   [AllDifferent(x[w] + y[w]) for w in range(nWeeks)],
   
   # each team plays at most two times in each period
   [Cardinality(x[:, p] + y[:, p], occurrences={t: range(1, 3) for t in range(nTeams)})
      for p in range(nPeriods)],

   # tag(symmetry-breaking)
   [
      # the match '0 versus t' (with t strictly greater than 0) appears at week t-1
      [Count(m[w], value=match_number(0, w + 1)) == 1 for w in range(nWeeks)],

      # the first week is set : 0 vs 1, 2 vs 3, 4 vs 5, etc.
      [m[0][p] == match_number(2 * p, 2 * p + 1) for p in range(nPeriods)]
   ]
)

if variant("dummy"):
   # xd[p] is the first team for the dummy match of period p  tag(dummy-week)
   xd = VarArray(size=nPeriods, dom=range(nTeams))

   # yd[p] is the second team for the dummy match of period p  tag(dummy-week)
   yd = VarArray(size=nPeriods, dom=range(nTeams))

   satisfy(
      # handling dummy week (variables and constraints)  tag(dummy-week)
      [
         # all teams are different in the dummy week
         AllDifferent(xd + yd),

         # each team plays two times in each period
         [Cardinality(x[:, p] + y[:, p] + [xd[p], yd[p]],
            occurrences={t: 2 for t in range(nTeams)}) for p in range(nPeriods)],
         
         # tag(symmetry-breaking)
         [xd[p] < yd[p] for p in range(nPeriods)]
      ]
   )    
\end{python}\end{boxpy}

This model involves $3+2$ arrays of variables and 5 types of constraints: \gb{Cardinality}, \gb{AllDifferent}, \gb{Count} (\gb{Exactly1}), \gb{Extension} and \gb{Intension}.
A series of 10 instances has been selected for the competition, for variant `dummy'.
For generating an \x3 instance (file), you can execute for example:
\begin{command}
python SportScheduling.py -data=10 -variant=dummy 
\end{command}

\subsection{Superpermutation}

\paragraph{Description (from Wikipedia).}

In combinatorial mathematics, a superpermutation on $n$ symbols is a string that contains each permutation of $n$ symbols as a substring.
While trivial superpermutations can simply be made up of every permutation listed together, superpermutations can also be shorter (except for the trivial case of $n = 1$) because overlap is allowed.
For instance, in the case of $n = 2$, the superpermutation 1221 contains all possible permutations (12 and 21),
but the shorter string 121 also contains both permutations.
It has been shown that for $1 \leq n \leq 5$, the smallest superpermutation on n symbols has length 1! + 2! + ... + n!.
The first four smallest superpermutations have respective lengths 1, 3, 9, and 33, forming the strings 1, 121, 123121321, and 123412314231243121342132413214321.
However, for n = 5, there are several smallest superpermutations having the length 153.
See \href{https://en.wikipedia.org/wiki/Stable_roommates_problem}{wikipedia.org}.

\paragraph{Data.}
Only one integer is required to specify a specific instance. Values of $n$ used for the instances in the competition are:
\begin{itemize}
\item 3, 4, 5 for main variant 
\item 3, 4, 5 for variant 'table' 
\end{itemize}

\paragraph{Model.}
The \p3 model, in a file `Superpermutation.py', used for the competition is: 

\begin{boxpy}\begin{python}
@\imp@

n = data  
m = sum(factorial(i) for i in range(1, n + 1))  # the length of the sequence
assert 2 <= n <= 5, "for the moment, the model is valid for n between 2 and 5"

permutations = list(permutations(list(range(1, n + 1))))
nPermutations = len(permutations)

# x[i] is the ith value of the sequence
x = VarArray(size=m, dom=range(1, n + 1))

if not variant():
   # p[j] is the index in the sequence of the first value of the jth permutation
   p = VarArray(size=nPermutations, dom=range(m))

   satisfy(
      # all permutations start at different indexes  tag(redundant-constraints)
      AllDifferent(p),

      # ensuring that each permutation occurs in the sequence
      [x[p[j] + k] == permutations[j][k] for k in range(n) for j in range(nPermutations)]
   )

elif variant("table"):
   nPatterns = m - n + 1  # a pattern is a possible subsequence of length n
   gap = nPatterns - nPermutations  # the gap corresponds to the flexibility we have

   T = [(i, *t) for i, t in enumerate(permutations)]
   T.extend((-1, *(v if k in (i, j) else ANY for k in range(n)))
         for v in range(n) for i, j in combinations(n, 2))

   # y[i] is the index of the permutation x[i:i+n] or -1 if this is not a permutation
   y = VarArray(size=nPatterns, dom=range(-1, nPermutations))

   satisfy(
      # identifying each pattern (subsequence of n values)
      [(y[i], x[i:i + n]) in T for i in range(nPatterns)],

      # ensuring that each permutation occurs in the sequence
      Cardinality(y, occurrences={-1: range(gap + 1)}
                        + {i: range(1, gap + 1) for i in range(nPermutations)})
   )

satisfy(
   # setting the first permutation  tag(symmetry-breaking)
   [x[i] == i + 1 for i in range(n)],

   # constraining a palindrome  tag(palindrome)
   [x[i] == x[-1 - i] for i in range(m // 2)]
)
\end{python}\end{boxpy}

Two variants of the problem are described here, involving 2 arrays of variables and different types of constraints: \gb{AllDifferent}, \gb{Element}, \gb{Cardinality}, \gb{Extension} and \gb{Intension}.
A series of $2\times3$ instances has been generated.

For generating an \x3 instance (file), you can execute for example:
\begin{command}
python Superpermutation.py -data=5 
python Superpermutation.py -data=5 -variant=table 
\end{command}

\section{COP}

\subsection{Aircraft Landing}

\paragraph{Description {\small (from Choco Tutorial)}.}
Given a set of planes and runways, the objective is to minimize the total (weighted) deviation from the target landing time for each plane.
There are costs associated with landing either earlier or later than a target landing time for each plane.
Each plane has to land on one of the runways within its predetermined time windows such that separation criteria between all pairs of planes are satisfied.
See \href{http://people.brunel.ac.uk/~mastjjb/jeb/orlib/airlandinfo.html}{OR-Library}.

\paragraph{Data.}
As an illustration of data specifying an instance of this problem, we have:

\begin{json}
{
  "n": 10,
  "times": [
    {"earliest": 129, "target": 155, "latest": 559},
    {"earliest": 195, "target": 258, "latest": 744},
    ...
  ], 
  "costs": [
    {"early_penalty": 1000, "late_penalty": 1000},
    {"early_penalty": 1000, "late_penalty": 1000},
    ...
  ],
  "separations": [
    [99999, 3, 15, 15, 15, 15, 15, 15, 15, 15],
    [3, 99999, 15, 15, 15, 15, 15, 15, 15, 15],
    ...
  ]
}
\end{json}

\paragraph{Model.}
The \p3 model, in a file `AircraftLanding.py', used for the competition is: 

\begin{boxpy}\begin{python}
@\imp@

nPlanes, times, costs, separations = data
earliest, target, latest = zip(*times)
early_penalties, late_penalties = zip(*costs)

# x[i] is the landing time of the ith plane
x = VarArray(size=nPlanes, dom=lambda i: range(earliest[i], latest[i] + 1))

# e[i] is the earliness of the ith plane
e = VarArray(size=nPlanes, dom=lambda i: range(target[i] - earliest[i] + 1))

# t[i] is the tardiness of the ith plane
t = VarArray(size=nPlanes, dom=lambda i: range(latest[i] - target[i] + 1))

satisfy(
   # planes must land at different times
   AllDifferent(x),

   # the separation time required between any two planes must be satisfied:
   [
      NoOverlap(
         origins=[x[i], x[j]],
         lengths=[separations[i][j], separations[j][i]]
      ) for i, j in combinations(nPlanes, 2)
   ]
)

if not variant():
   satisfy(
      # computing earlinesses of planes
      [e[i] == max(0, target[i] - x[i]) for i in range(nPlanes)],

      # computing tardinesses of planes
      [t[i] == max(0, x[i] - target[i]) for i in range(nPlanes)],
   )
elif variant("table"):
   satisfy(
      # computing earlinesses and tardinesses of planes
     (x[i], e[i], t[i]) in {(v, max(0, target[i] - v), max(0, v - target[i]))
        for v in range(earliest[i], latest[i] + 1)} for i in range(nPlanes)
   )

minimize(
   # minimizing the deviation cost
   e * early_penalties + t * late_penalties
)
\end{python}\end{boxpy}

Two variants of the problem are described here (but only the variant 'table' is used for the competition), involving 3 arrays of variables and different types of constraints: \gb{AllDifferent}, \gb{NoOverlap} and \gb{Extension}.
A series of 13 instances (coming from \href{http://people.brunel.ac.uk/~mastjjb/jeb/orlib/airlandinfo.html}{OR-Library}) has been selected.

For generating an \x3 instance (file), you can execute for example:
\begin{command}
python AircraftLanding.py -variant=table -data=airland01.txt
                          -parser=AircraftLanding_Parser.py 
\end{command}
where `airland01.txt' is a data file and `AircraftLanding\_Parser.py' is a parser (i.e., a Python file allowing us to load data that are not directly given in JSON format).
Note that for saving data in JSON files, you can add the option `-export' (or `-dataexport').

\subsection{Clock Triplets}

\paragraph{Description.}
Martin Gardner presented this problem:
\begin{quote}
  Now for a curious little combinatorial puzzle involving the twelve numbers on the face of a clock.
  Can you rearrange the numbers (keeping them in a circle) so no triplet of adjacent numbers has a sum higher than 21?
  This is the smallest value that the highest sum of a triplet can have.
\end{quote}
See \href{http://www.f1compiler.com/samples/Dean%20Clark%27s%20Problem.f1.html}{f1compiler.com}
The problem here is given in a general form.
  
\paragraph{Data.}
Only two integers are required to specify a specific instance. Values of $r$ and $n$ used for the instances in the competition are:
\begin{quote}
(3,12), (5,15), (7,15), (7,20), (10,15), (10,20), (15,20), (15,30), (20,25), (20,35)
\end{quote}

\paragraph{Model.}
The \p3 model, in a file `ClockTriplet.py', used for the competition is: 

\begin{boxpy}\begin{python}
@\imp@

r, n = data 

# x[i] is the ith number in the circle
x = VarArray(size=n, dom=range(1, n + 1))

# z is the minimal value such that any (circular) subsequence of x  of size r is <= z
z = Var(range(sum(n - v for v in range(r)) + 1))

satisfy(
   # a permutation is required
   AllDifferent(x),

   # any subsequence of size r must be less than or equal to z
   [Sum(x[j] for j in [(i + k) % n for k in range(r)]) <= z for i in range(n)],

   # tag(symmetry-breaking)
   [x[0] == 1, x[1] < x[-1]]
)

minimize(
   z
)
\end{python}\end{boxpy} 

This model involves 1 array of variables, a stand-alone variable,  and 3 types of constraints: \gb{AllDifferent}, \gb{Sum} and \gb{intension}.
A series of 10 instances has been selected for the competition.
For generating an \x3 instance (file), you can execute for example:
\begin{command}
python ClockTriplet.py -data=[7,20]
\end{command}

\subsection{Coins Grid}

\paragraph{Description.}

The problem, from Tony Hurlimann's working paper called `A coin puzzle: SVOR-contest 2007', is defined as follows.
In a quadratic grid (or a larger chessboard) with $n \times n$ cells, one should place $c$ coins in such a way that the following conditions are fulfilled:
\begin{enumerate}
\item In each row exactly $c$ coins must be placed.
\item In each column exactly $c$ coins must be placed.
\item The sum of the quadratic horizontal distance from the main diagonal of all cells containing a coin must be as small as possible.
\item In each cell at most one coin can be placed.
\end{enumerate}
The problem is initially illustrated with $n=31$ and $c=14$.
  
\paragraph{Data.}
Only two integers are required to specify a specific instance. Values of $n$ and $c$ used for the instances in the competition are:
\begin{quote}
(8,4), (10,5), (12,6), (14,7), (16,8), (19,9), (22,10), (25,11), (28,12), (31,14)
\end{quote}

\paragraph{Model.}
The \p3 model, in a file `CoinsGrid.py', used for the competition is: 

\begin{boxpy}\begin{python}
@\imp@

n, c = data

# x[i][j] is 1 if a coin is placed at row i and column j
x = VarArray(size=[n, n], dom={0, 1})

satisfy(
   [Sum(x[i]) == c for i in range(n)],

   [Sum(x[:, j]) == c for j in range(n)]
)

minimize(
   Sum(x[i][j] * abs(i - j) ** 2 for i in range(n) for j in range(n))
)
\end{python}\end{boxpy} 

This model involves 1 array of variables, and some constraints \gb{Sum}.
A series of 10 instances has been selected for the competition.
For generating an \x3 instance (file), you can execute for example:
\begin{command}
python CoinsGrid.py -data=[31,14]
\end{command}

\subsection{CVRP}

This is Problem \href{https://www.csplib.org/Problems/prob086}{086} on CSPLib, called Capacitated Vehicle Routing Problem.

\paragraph{Description.}
The capacitated vehicle routing problem (CVRP) is a VRP in which vehicles with limited carrying capacity need to pick up or deliver items at various locations.
The items have a quantity, such as weight or volume, and the vehicles have a maximum capacity that they can carry.
See \href{http://vrp.galgos.inf.puc-rio.br/index.php/en/}{CVRPLIB}.

\paragraph{Data.}
As an illustration of data specifying an instance of this problem, we have:

\begin{json}
{
  "n": 32,
  "capacity": 100,
  "demands": [0,19,...],
  "distances":[
    [0,35,...],
    [35,0,...],
    ...
  ]
}
\end{json}

\paragraph{Model.}
The \p3 model, in a file `CVRP.py', used for the competition is: 

\begin{boxpy}\begin{python}
@\imp@

nNodes, capacity, demands, distances = data
nVehicles = nNodes // 4  # hard coding, which can be at least used for Set A (Augerat, 1995)

def max_tour():
   t = sorted(demands)
   i, s = 1, 0
   while i < nNodes and s < capacity:
      s += t[i]
      i += 1
   return i - 2

nSteps = max_tour()
n0s = nVehicles * nSteps - nNodes + 1

# c[i][j] is the jth customer (step) during the tour of the ith vehicle
c = VarArray(size=[nVehicles, nSteps], dom=range(nNodes))

# d[i][j] is the demand of the jth customer during the tour of the ith vehicle
d = VarArray(size=[nVehicles, nSteps], dom=demands)

satisfy(
   AllDifferent(c, excepting=0),

   # ensuring that all demands are satisfied
   Cardinality(c, occurrences={i: 1 if i > 0 else n0s for i in range(nNodes)}), 

   # no holes permitted during tours
   [
     If(
        c[i][j] == 0,
        Then=c[i][j+1] == 0
     ) for i in range(nVehicles) for j in range(nSteps - 1)
   ],
   
   # computing the collected demands
   [demands[c[i][j]] == d[i][j] for i in range(nVehicles) for j in range(nSteps)],
   
   # not exceeding the capacity of each vehicle
   [Sum(d[i]) <= capacity for i in range(nVehicles)],
   
   # tag(symmetry-breaking)
   Decreasing(c[:, 0])
)

minimize(
   # minimizing the total traveled distance by vehicles
     Sum(distances[0][c[i][0]] for i in range(nVehicles))
   + Sum(distances[c[i][j]][c[i][j+1]] for i in range(nVehicles) for j in range(nSteps - 1))
   + Sum(distances[c[i][-1]][0] for i in range(nVehicles))
)
\end{python}\end{boxpy}

This problem involves 2 arrays of variables and 5 types of constraints: \gb{AllDifferent}, \gb{Cardinality}, \gb{Sum}, \gb{Decreasing} and \gb{Intension}.
A series of 10 instances has been selected for the competition (coming from \href{http://vrp.galgos.inf.puc-rio.br/index.php/en/}{CVRPLIB}).
For generating an \x3 instance (file), you can execute for example:
\begin{command}
python CVRP.py -data=A-n32-k5.json
\end{command}

\subsection{Cyclic Bandwith}

\paragraph{Description.}
The Cyclic Bandwidth problem is a graph embedding problem.
It was first stated by Leung et al. in 1984 in relation with the design of a ring interconnection network.
Their aim was to find an arrangement on a cycle for a set of computers with a known communication pattern given by a graph, in such a way that every message sent can arrive at its destination in at most k steps.
The CB problem arises also in other important application areas like VLSI designs, data structure representations, code design and interconnection networks for parallel computer systems.
See \href{https://www.tamps.cinvestav.mx/~ertello/cbmp.php}{E. Rodriguez-Tello's page}.

\paragraph{Data.}
As an illustration of data specifying an instance of this problem, we have:

\begin{json}
{
  "n": 104,
  "edges": [[30, 101], [101, 43], ... ]
}
\end{json}

\paragraph{Model.}
The \p3 model, in a file `CyclicBandwith.py', used for the competition is: 

\begin{boxpy}\begin{python}
@\imp@

n, edges = data 

# x[i] is the label of the ith node
x = VarArray(size=n, dom=range(n))

satisfy(
   AllDifferent(x)
)

minimize(
   Maximum(min(abs(x[i] - x[j]), n - abs(x[i] - x[j])) for i, j in edges)
)
\end{python}\end{boxpy}

This problem involves 1 array of variables, 1 constraint \gb{AllDifferent} and a complex objective expression.
A series of 10 instances has been selected for the competition (coming from \href{https://www.tamps.cinvestav.mx/~ertello/cbmp.php}{E. Rodriguez-Tello's page}).
For generating an \x3 instance (file), you can execute for example:
\begin{command}
python CyclicBandwith.py -data=path100.rnd -parser=CyclicBandwith_Parser.py
\end{command}
where `path100.rnd' is a data file and `CyclicBandwith\_Parser.py' is a parser (i.e., a Python file allowing us to load data that are not directly given in JSON format).
Note that for saving data in JSON files, you can add the option `-export' (or `-dataexport').

\subsection{DC}

A series of 26 instances linked to DC (Differential Cryptanalysis) has been generated (independently of the library \p3), and submitted to the 2022 competition by Fran\c cois Delobel.
These instances have been generated with Tagada \cite{tagada} and are about finding truncated differential features for symmetric encryption algorithms.
More specifically, all instances consist in finding the smallest number of active S-boxes in a truncated differential characteristic (problem called Step1-opt in the CP paper).
The encryption algorithms considered are Midori, Rijndael and Skinny. There are multiple instances for each cipher, with an increasing number $r$ of rounds.

\subsection{Echelon Stock}

This is Problem \href{https://www.csplib.org/Problems/prob040}{040} on CSPLib, called Distribution Problem with Wagner-Whitin Costs.

\paragraph{Description {\small (excerpt from CSPLib)}.}

A basic distribution problem is described as follows.
Given:
\begin{itemize}
\item a supply chain structure of stocking points divided into levels
\item a holding cost per unit of inventory at each stocking point, where it is assumed that a parent has lower holding cost than any of its children
\item a procurement cost per stocking point (per order, not per unit of inventory received)
\item a number of periods
\item a demand for each leaf at each period
\end{itemize}
find an optimal ordering policy: i.e. a decision as to how much to order at each stocking point at each time period that minimises cost.
The Wagner-Whitin form of the problem assumes that the holding costs and procurement costs are constant, and that the demands are known for the entire planning horizon. Furthermore, the stocking points have no maximum capacity and the starting inventory is 0.

\paragraph{Data.}
As an illustration of data specifying an instance of this problem, we have:

\begin{json}
{
  "children": [[], [], [], [0, 1], [2], [3, 4]],
  "hcosts": [3, 3, 3, 2, 2, 1],
  "pcosts": [1000, 1000, 1000, 1000, 1000, 1000],
  "demands": [[100, 100, 100, 100, 100, 100, 100, 100, 100, 100, 100, 100],
    [50, 200, 50, 50, 200, 250, 250, 100, 150, 150, 50, 200],
    [250, 50, 350, 50, 250, 50, 250, 50, 350, 100, 50, 50]]
}
\end{json}

\paragraph{Model.}
The \p3 model, in a file `EchelonStock2.py', used for the competition is: 

\begin{boxpy}\begin{python}
@\imp@

children, hcosts, pcosts, demands = data
n, nPeriods, nLeaves = len(children), len(demands[0]), len(demands)

# below, some simplifications
gcd = reduce(gcd, {v for row in demands for v in row})
demands = [[row[t] // gcd for t in range(nPeriods)] for row in demands]
hcosts = [hcosts[i] * gcd for i in range(n)]

sum_dmds, all_dmds = [], []
for i in range(n):
   if i < nLeaves:
      sum_dmds.append(sum(demands[i]))
      all_dmds.append([sum(demands[i][t:]) for t in range(nPeriods)])
   else:
      sum_dmds.append(sum(sum_dmds[j] for j in children[i]))
      all_dmds.append([sum(all_dmds[j][t] for j in children[i]) for t in range(nPeriods)])

def ratio1(i, coeff=1):
   parent = next(j for j in range(n) if i in children[j])  
   return floor(pcosts[i] // (coeff * (hcosts[i] - hcosts[parent])))

def ratio2(i, t_inf):
   return min(sum(demands[i][t_inf: t_sup + 1]) + ratio1(i, t_sup - t_inf + 1)
      for t_sup in range(t_inf, nPeriods))

def domain_x(i, t):  # ratio2 from IC4, and all_dmds from IC6a
   return range(min(all_dmds[i][t], ratio2(i, t)) + 1) if i < nLeaves
           else range(all_dmds[i][t] + 1)

def domain_y(i, t):  # {0} from IC1, ratio1 from IC3 and all_dmds from IC6b 
   return {0} if t == nPeriods - 1 else range(min(all_dmds[i][t + 1], ratio1(i)) + 1)
           if i < n - 1 else range(all_dmds[i][t + 1] + 1)

# x[i][t] is the amount ordered at node i at period (time) t
x = VarArray(size=[n, nPeriods], dom=domain_x)

# y[i][t] is the amount stocked at node i at the end of period t
y = VarArray(size=[n, nPeriods], dom=domain_y)

satisfy(
   [y[i][0] == x[i][0] - demands[i][0] for i in range(nLeaves)],

   [y[i][t] == x[i][t] + y[i][t - 1] - demands[i][t] for i in range(nLeaves)
      for t in range(1, nPeriods)],

   [y[i][0] == x[i][0] - Sum(x[j][0] for j in children[i]) for i in range(nLeaves, n)],

   [y[i][t] == x[i][t] + y[i][t - 1] - Sum(x[j][t] for j in children[i])
      for i in range(nLeaves, n) for t in range(1, nPeriods)],


   # IC2
   [(x[i][t] == 0) | disjunction(x[j][t] > 0 for j in children[i])
      for i in range(nLeaves, n) for t in range(nPeriods)],

   # IC5
   [(y[i][t - 1] == 0) | (x[i][t] == 0) for i in range(n) for t in range(1, nPeriods)],

   # tag(redundant-constraints)
   [Sum(x[i]) == sumDemands[i] for i in range(n)],

   [y[i][t - 1] + Sum(x[i][t:]) == all_dmds[i][t] for i in range(nLeaves)
      for t in range(1, nPeriods)]
)

minimize(
     Sum(hcosts[i] * y[i][t] for i in range(n) for t in range(nPeriods))
   + Sum(pcosts[i] * (x[i][t] > 0) for i in range(n) for t in range(nPeriods))
)
\end{python}\end{boxpy}

This problem involves 2 arrays of variables, 2 types of constraints \gb{Sum} and \gb{Intension}, and a complex objective expression.
A series of 10 instances has been selected for the competition (coming from \href{https://www.csplib.org/Problems/prob040}{CSPLib}).
For generating an \x3 instance (file), you can execute for example:

\begin{command}
python EchelonStock2.py -data=A01.txt -parser=EchelonStock_Parser.py
\end{command}

where `A01.txt' is a data file and `EchelonStock\_Parser.py' is a parser (i.e., a Python file allowing us to load data that are not directly given in JSON format).
Note that for saving data in JSON files, you can add the option `-export' (or `-dataexport').

\subsection{Filters}

\paragraph{Description.}
This problem is about optimizing the scheduling of filter operations, commonly used in High-Level Synthesis.
This problem/model has been originally written by Krzysztof Kuchcinski for the 2010, 2012, 2013 and 2016 Minizinc competitions.
See also the models in \href{https://github.com/radsz/jacop/tree/develop/src/main/java/org/jacop/examples/fd/filters}{JaCop}.

\paragraph{Data.}
As an illustration of data specifying an instance of this problem, we have:

\begin{json}
{
  "del_add": 1,
  "del_mul": 2,
  "number_add": 2,
  "number_mul": 3,
  "last": [40, 41, 42, 43, 44, 45, 46, 47],
  "add": [0, 1, 2, 3, 4, ...],
  "dependencies": [[8, 16, 17], [8, 19, 20], ...]
}
\end{json}

\paragraph{Model.}
The \p3 model, in a file `Filters.py', used for the competition is: 

\begin{boxpy}\begin{python}
@\imp@

d_add, d_mul, n_add, n_mul, last, add, dependencies = data
nOperations = len(dependencies)

d = [d_add if i in add else d_mul for i in range(nOperations)]
mul = [i for i in range(nOperations) if i not in add]

# t[i] is the starting time of the ith operation
t = VarArray(size=nOperations, dom=range(101))

# r[i] is the (index of the) operator used for the ith operation
r = VarArray(size=nOperations, dom=lambda i: range(1, 1 + (n_add if i in add else n_mul)))

satisfy(
   # respecting dependencies
   [t[i] + d[i] <= t[j] for i in range(nOperations) for j in dependencies[i]],

   # no overlap concerning add operations
   NoOverlap(origins=[(t[i], r[i]) for i in add], lengths=[(d_add, 1) for i in add]),

   # no overlap concerning mul operations
   NoOverlap(origins=[(t[i], r[i]) for i in mul], lengths=[(d_mul, 1) for i in mul])
)

minimize(
   # minimizing the ending time of last operations
   Maximum(t[i] + d[i] for i in last)
)
\end{python}\end{boxpy}

This problem involves 2 arrays of variables and two types of constraints: \gb{NoOverlap} and \gb{Intension}.
A series of 8 instances has been selected for the competition (coming from data generated by K. Kuchcinski who submitted them to several Minizinc competitions).
For generating an \x3 instance (file), you can execute for example:
\begin{command}
python Filters.py -data=dct_1_1.dzn -parser=Filters_ParserZ.py
\end{command}
where `dct\_1\_1.dzn' is a data file and `Filters\_ParserZ.py' is a parser (i.e., a Python file allowing us to load data that are not directly given in JSON format).
Note that for saving data in JSON files, you can add the option `-export' (or `-dataexport').

\subsection{Itemset Mining}

\paragraph{Description.}
A model for the following problem has been originally written by Tias Guns for the 2011, 2012 and 2013 Minizinc competitions.
A traditional task in machine learning is the task of concept learning.
Given a dataset of positive and negative examples, the aim is here to find a formula in disjunctive normal form which characterizes the positive examples  as accurately as possible.
In this challenge this task is modeled as a discrete constraint optimization problem; the aim is to find a formula which is as accurate as possible.
 
The model is based on the link between DNF formulas and pattern sets in the data mining literature. It represents the formula as a set of itemsets, and imposes constraints on both the itemsets and the set of itemsets.
It is based on the 'Constraint Programming for Itemset Mining' framework called \href{https://dtai.cs.kuleuven.be/CP4IM}{CP4IM}.
See also \cite{GNR_itemset}.

\paragraph{Data.}
As an illustration of data specifying a toy instance of this problem, we have:

\begin{json}
{
  "nItems": 7,
  "pos": [[1, 3], [2, 4, 6]],
  "neg": [[0, 2, 5], [1, 2, 6], [2, 3, 5],
  "k": 2
}
\end{json}

\paragraph{Model.}
The \p3 model, in a file `ItemsetMining.py', used for the competition is: 

\begin{boxpy}\begin{python}
@\imp@

nItems, positiveExamples, negativeExamples, nSets = data  # nSets is the original k
nPos, nNeg = len(positiveExamples), len(negativeExamples)
assert nSets in (1, 2)

# precomputing three auxiliary complementary sets
posC = [[i for i in range(nItems) if i not in t] for t in positiveExamples]
negC = [[i for i in range(nItems) if i not in t] for t in negativeExamples]
itmC = [[j for j in range(nPos) if i not in positiveExamples[j]] for i in range(nItems)]

if nSets == 1:
   # x[i] is 1 if the ith item is selected
   x = VarArray(size=nItems, dom={0, 1})

   # tp[j] is 1 if the jth positive example is a true positive
   tp = VarArray(size=nPos, dom={0, 1})

   # tn[j] is 1 if the jth negative example is a true negative
   tn = VarArray(size=nNeg, dom={0, 1})


   satisfy(
      # computing true positives
      [tp[j] == NotExist(x[t]) for j, t in enumerate(posC) if len(t) > 0],

      # computing true negatives
      [tn[j] == NotExist(x[t]) for j, t in enumerate(negC) if len(t) > 0],

      # computing selected items
      [x[i] == NotExist(tp[t]) for i, t in enumerate(itmC) if len(t) > 0]
   )

   maximize(
      # maximizing correct discrimination
      Sum(tp) - Sum(tn)
   )

else:  # nSets = 2

   # x[k][i] is 1 if the ith item is selected in the kth set
   x = VarArray(size=[nSets, nItems], dom={0, 1})

   # tp[k][j] is 1 if the jth positive example is a true positive for the kth set
   tp = VarArray(size=[nSets, nPos], dom={0, 1})
   
   # tp[k][j] is 1 if the jth positive example is a true positive for the kth set
   tn = VarArray(size=[nSets, nNeg], dom={0, 1})
   
   jtp = VarArray(size=nPos, dom={0, 1})
   
   jtn = VarArray(size=nNeg, dom={0, 1})
   
   satisfy(
      # computing true positives
      [tp[k][j] == NotExist(x[k][t]) for k in range(nSets) for j, t in enumerate(posC)
         if len(t) > 0],

      # computing true negatives
      [tn[k][j] == NotExist(x[k][t]) for k in range(nSets) for j, t in enumerate(negC)
         if len(t) > 0],
      
      # computing selected items
      [x[k][i] == NotExist(tp[k][t]) for k in range(nSets) for i, t in enumerate(itmC)
         if len(t) > 0],
      
      # computing joint true positives
      [jtp[t] == Exist(tp[:, t]) for t in range(nPos)],
      
      # computing joint true negatives
      [jtn[t] == Exist(tn[:, t]) for t in range(nNeg)],
      
      # tag(symmetry-breaking)
      [
        LexIncreasing(tp, strict=True),
        LexIncreasing(tn, strict=True)
      ]
   )

   maximize(
      # maximizing joint correct discrimination
      Sum(jtp) - Sum(jtn)
   )
\end{python}\end{boxpy}

This model involves 3 and 5 arrays of variables (depending on the value of $k$) and several types of constraints: \gb{Count}, \gb{Intension}, \gb{LexIncreasing} and \gb{Sum}.
A series of 15 instances has been selected for the competition (coming from data generated by T. Guns who submitted them to Minizinc competitions).
For generating an \x3 instance (file), you can execute for example:
\begin{command}
python ItemsetMining.py -data=audiology-k2.dzn -parser=ItemsetMining_ParserZ.py
\end{command}
where `audiology-k2.dzn' is a data file and `ItemsetMining\_ParserZ.py' is a parser (i.e., a Python file allowing us to load data that are not directly given in JSON format).
Note that for saving data in JSON files, you can add the option `-export' (or `-dataexport').

\subsection{Multi-Agent Path Finding}

\paragraph{Description.}
A model for the following problem has been originally written by Neng-Fa Zhou in Picat (and translated to Minizinc by Hakan Kjellerstrand for the 2017 Minizinc competition).
\begin{quote}
The multi-agent pathfinding (MAPF) problem amounts to finding a plan for agents to move within a graph from their starting locations to their destinations, such that no agents collide with each other at any time.
MAPF can be solved suboptimally in polynomial time [14], but finding an optimal solution is NP-hard for common optimization criteria
\end{quote}

See also \cite{mapf}.

\paragraph{Data.}
As an illustration of data specifying an instance of this problem, we have:

\begin{json}
{
  "agents": [[104, 31], [60, 56], [125, 211], ...],
  "horizon": 113,
  "neighbors": [[0, 15], [1, 17, 2], [2, 18, 1, 3], ...]
}
\end{json}

\paragraph{Model.}
The \p3 model, in file `MultiAgentPathFinding.py', used for the competition is: 

\begin{boxpy}\begin{python}
@\imp@

agents, horizon, neighbors, = data
src, dst = zip(*agents)
nAgents, nNodes = len(agents), len(neighbors)

if variant("table"):

   T = [(i, v) for i, t in enumerate(neighbors) for v in t]

   # x[t][a] is the node where is the agent a at time t
   x = VarArray(size=[horizon + 1, nAgents], dom=range(nNodes))

   # e[a] is the time when the agent a arrives at its destination
   e = VarArray(size=nAgents, dom=range(horizon + 1))
   
   satisfy(
      # agents must occupy different node at any time
      [AllDifferent(x[t]) for t in range(horizon + 1)],

      # agents at their destinations stays there
      [
         If(
            x[t][a] == dst[a],
            Then= x[t + 1][a] == dst[a]
         ) for t in range(horizon) for a in range(nAgents)
      ],
      
      # agents can only move to connected nodes
      [(x[t][a], x[t + 1][a]) in T for t in range(horizon) for a in range(nAgents)],

      # setting agents at their initial positions
      [x[0][a] == src[a] for a in range(nAgents)],

      # setting agents at their final positions
      [x[-1][a] == dst[a] for a in range(nAgents)],

      # computing end times of agents
      [
        (e[a] == t + 1) == both(
           x[t][a] != dst[a],
           x[t + 1][a] == dst[a])
        ) for t in range(horizon) for a in range(nAgents)
      ]
   )

   if subvariant("mks"):
      minimize(
         Maximum(e)
      )
   else:
      minimize(
         Sum(e) + nAgents
      )
\end{python}\end{boxpy}

This problem involves 2 arrays of variables and 3 types of constraints: \gb{AllDifferent}, \gb{Extension} and \gb{Intension}.
A series of $2 \times 10$ instances has been selected for the competition (coming from data gently given by N.-F. Zhou).
For generating an \x3 instance (file), you can execute for example:
\begin{command}
python MultiAgentPathFinding.py -variant=table -data=g16_p10_a05.pi 
                                -parser=MultiAgentPathFinding_ParserPicat.py
python MultiAgentPathFinding.py -variant=table-mks -data=g16_p10_a05.pi 
                                -parser=MultiAgentPathFinding_ParserPicat.py
\end{command}
where `g16\_p10\_a05.pi' is a data file and `MultiAgentPathFinding\_ParserPicat.py' is a parser (i.e., a Python file allowing us to load data that are not directly given in JSON format).
Note that for saving data in JSON files, you can add the option `-export' (or `-dataexport').

\subsection{Nurse Rostering}

This is a realistic employee shift scheduling Problem (see, for example, \cite{MBL_solving}).

\paragraph{Description.}
The description is rather complex. Hence, we refer the reader to:\\ \href{http://www.schedulingbenchmarks.org/nrp/}{http://www.schedulingbenchmarks.org/nrp/}.

\paragraph{Data.}
As an illustration of data specifying an instance of this problem, we have:

\begin{json}
{
  "nDays": 14,
  "shifts": [ { "id": "D", "length": 480, "forbiddenFollowingShifts": "null" } ],
  "staffs": [
    { "id": "A",
      "maxShifts": [14],
      "minTotalMinutes": 3360,"maxTotalMinutes": 4320,
      "minConsecutiveShifts": 2,"maxConsecutiveShifts": 5,
      "minConsecutiveDaysOff": 2,
      "maxWeekends": 1, "daysOff": [0],
      "onRequests": [
        { "day": 2, "shift": "D", "weight": 2 },
        { "day": 3, "shift": "D", "weight": 2}
      ],
      "offRequests": "null"
    },
    ...
  ],
  "covers": [
     [ { "requirement": 3, "weightIfUnder": 100, "weightIfOver": 1 } ],
     [ { "requirement": 5, "weightIfUnder": 100, "weightIfOver": 1 } ],
     ...
  ]
}
\end{json}

\paragraph{Model.}
The \p3 model, in a file `NurseRostering.py', used for the competition is: 

\begin{boxpy}\begin{python}
@\imp@

nDays, shifts, staffs, covers = data

if shifts[-1].id != "_off":  # if not present, we add first a dummy 'off' shift
   shifts.append(shifts[0].__class__("_off", 0, None))  # with a named tuple of the same class
OFF = len(shifts) - 1  # value for _off
lengths = [shift.length for shift in shifts]
nWeeks, nShifts, nStaffs = nDays // 7, len(shifts), len(staffs)

on_r = [[next((r for r in staff.onRequests if r.day == day), None) if staff.onRequests
          else None for day in range(nDays)] for staff in staffs]
off_r = [[next((r for r in staff.offRequests if r.day == day), None) if staff.offRequests
          else None for day in range(nDays)] for staff in staffs]

# kmin, kmax, kday for minConsecutiveShifts, maxConsecutiveShifts, minConsecutiveDaysOff
_, maxShifts, minTimes, maxTimes, kmin, kmax, kday, maxWeekends, daysOff, _, _ = zip(*staffs)

sp = {shifts[i].id: i for i in range(nShifts)}  # position of shifts in the list 'shifts'
T = {(sp[s1.id], sp[s2]) for s1 in shifts if s1.forbiddenFollowingShifts
          for s2 in s1.forbiddenFollowingShifts}  # rotation

def costs(day, shift):
   if shift == OFF: return [0] * (nStaffs + 1)
   r, wu, wo = covers[day][shift]
   return [abs(r - i) * (wu if i <= r else wo) for i in range(nStaffs + 1)]

def automaton(k, for_shifts):  # automaton_min_consecutive
   q = Automaton.q  # for building state names
   range_off = range(nShifts - 1, nShifts)  # a range with only one value (off)
   range_others = range(nShifts - 1)        # a range with all other values
   r1, r2 = (range_off,range_others) if for_shifts else (range_others, range_off)
   t = [(q(0), r1, q(1)), (q(0), r2, q(k + 1)), (q(1), r1, q(k + 1))]
   t.extend((q(i), r2, q(i + 1)) for i in range(1, k + 1))
   t.append((q(k + 1), range(nShifts), q(k + 1)))
   return Automaton(start=q(0), final=q(k + 1), transitions=t)

# x[d][p] is the shift at day d for person p (value 'OFF' denotes off)
x = VarArray(size=[nDays, nStaffs], dom=range(nShifts))

# nd[p][s] is the number of days such that person p works with shift s
nd = VarArray(size=[nStaffs, nShifts],
                dom=lambda p, s: range((nDays if s == OFF else maxShifts[p][s]) + 1))

# np[d][s] is the number of persons working on day d with shift s
np = VarArray(size=[nDays, nShifts], dom=range(nStaffs + 1))

# wk[p][w] is 1 iff the week-end w is worked by person p
wk = VarArray(size=[nStaffs, nWeeks], dom={0, 1})


# cn[p][d] is the cost of not satisfying the on-request (if it exists) of person p on day d
cn = VarArray(size=[nStaffs, nDays],
                dom=lambda p, d: {0, on_r[p][d].weight} if on_r[p][d] else None)

# cf[p][d] is the cost of not satisfying the off-request (if it exists) of person p on day d
cf = VarArray(size=[nStaffs, nDays],
                dom=lambda p, d: {0, off_r[p][d].weight} if off_r[p][d] else None)

# cc[d][s] is the cost of not satisfying cover for shift s on day d
cc = VarArray(size=[nDays, nShifts], dom=lambda d, s: costs(d, s))

satisfy(
   # days off for staff
   [x[d][p] == OFF for d in range(nDays) for p in range(nStaffs) if d in daysOff[p]],

   # computing number of days
   [Count(x[:, p], value=s) == nd[p][s] for p in range(nStaffs) for s in range(nShifts)],
   
   # computing number of persons
   [Count(x[d], value=s) == np[d][s] for d in range(nDays) for s in range(nShifts)],
   

   # computing worked weekends
   [
      (
         If(x[w * 7 + 5][p] != OFF, Then=wk[p][w]),
         If(x[w * 7 + 6][p] != OFF, Then=wk[p][w])
      ) for p in range(nStaffs) for w in range(nWeeks)
   ],

   # rotation shifts
   [(x[i][p], x[i + 1][p]) not in T for i in range(nDays - 1) for p in range(nStaffs)]
      if len(table) > 0 else None,
   
   # maximum number of worked week-ends
   [Sum(wk[p]) <= maxWeekends[p] for p in range(nStaffs)],
   
   # minimum and maximum number of total worked minutes
   [nd[p] * lengths in range(minTimes[p], maxTimes[p] + 1) for p in range(nStaffs)],
   
   # maximum consecutive worked shifts
   [Count(x[i:i + kmax[p] + 1, p], value=OFF) >= 1 for p in range(nStaffs)
      for i in range(nDays - kmax[p])],
   
   # minimum consecutive worked shifts
   [x[i: i + kmin[p] + 1, p] in automaton(kmin[p], True) for p in range(nStaffs)
      for i in range(nDays - kmin[p])],
   
   # managing off days on schedule ends
   [
      (
         If(x[0][p] != OFF, Then=x[i][p] != OFF),
         If(x[-1][p] != OFF, Then=x[-1 - i][p] != OFF)
      ) for p in range(nStaffs) if kmin[p] > 1 for i in range(1, kmin[p])
   ],

   
   # minimum consecutive days off
   [x[i: i + kday[p] + 1, p] in automaton(kday[p], False) for p in range(nStaffs)
      for i in range(nDays - kday[p])],
   
   # cost of not satisfying on requests
   [(x[d][p] == sp[on_r[p][d].shift]) == (cn[p][d] == 0) for p in range(nStaffs)
      for d in range(nDays) if on_r[p][d]],
   
   # cost of not satisfying off requests
   [(x[d][p] == sp[off_r[p][d].shift]) == (cf[p][d] != 0) for p in range(nStaffs)
      for d in range(nDays) if off_r[p][d]],
   
   # cost of under or over covering
   [(np[d][s], cc[d][s]) in enumerate(costs(d, s)) for d in range(nDays)
      for s in range(nShifts)]
)

minimize(
   Sum(cn) + Sum(cf) + Sum(cc)
)
\end{python}\end{boxpy}

This model involves 7 arrays of variables and 5 types of constraints: \gb{Regular}, \gb{Count}, \gb{Sum}, \gb{Intension} and \gb{Extension}.
Note that a series of 20 instances has been selected from \href{http://www.schedulingbenchmarks.org/}{http://www.schedulingbenchmarks.org}.
For generating an \x3 instance (file), you can execute for example:
\begin{command}
python NurseRostering.py -data=01 -parser=NurseRostering_Parser.py
\end{command}
where `01' is a data file and `NurseRostering\_Parser.py' is a parser (i.e., a Python file allowing us to load data that are not directly given in JSON format).
Note that for saving data in JSON files, you can add the option `-export' (or `-dataexport').  

\medskip
{\em Important:} The series of instances, used for the 2022 competition comes from the 2018 competition, and compiling instances from the \p3 model above may produce slighlty different files.

\subsection{Nursing Workload}

This is Problem \href{https://www.csplib.org/Problems/prob069}{069} on CSPLib, called Balanced Nursing Workload Problem.

\paragraph{Description {\small (excerpt from CSPLib)}.}
Given a set of patients distributed in a number of hospital zones and an available nursing staff, one must assign a nurse to each patient in such a way that the work is distributed evenly between nurses.
Each patient is assigned an acuity level corresponding to the amount of care he requires; the workload of a nurse is defined as the sum of the acuities of the patients he cares for.
A nurse can only work in one zone and there are retrictions both on the number of patients assigned to a nurse and on the corresponding workload.
We balance the workloads by minimizing their standard deviation.

\paragraph{Data.}
As an illustration of data specifying an instance of this problem, we have:

\begin{json}
{
  "nNurses": 11,
  "minPatientsPerNurse": 1,
  "maxPatientsPerNurse": 3,
  "maxWorkloadPerNurse": 105,
  "demands": [[59, 57, 50, 44, 42, 40, 39, 39, 33, 33, 32, 27, 26, 22, 20, 17, 11], [49, 47, 39, 39, 38, 30, 30, 28, 27, 15, 14]]
}
\end{json}

\paragraph{Model.}
The \p3 model, in a file `NursingWorkload.py', used for the competition is: 

\begin{boxpy}\begin{python}
@\imp@

nNurses, minPatientsPerNurse, maxPatientsPerNurse, maxWorkloadPerNurse, demands = data
patients = [(i, demand) for i, t in enumerate(demands) for demand in t]
nPatients, nZones = len(patients), len(demands)

lb = sum(sorted(dem for i, t in enumerate(demands) for dem in t)[:minPatientsPerNurse])

# p[i] is the nurse assigned to the ith patient
p = VarArray(size=nPatients, dom=range(nNurses))

# w[k] is the workload of the kth nurse
w = VarArray(size=nNurses, dom=range(lb, maxWorkloadPerNurse + 1))

satisfy(
   Cardinality(p, occurrences={k: range(minPatientsPerNurse, maxPatientsPerNurse + 1)
      for k in range(nNurses)}),

   [p[i] != p[j] for i, j in combinations(nPatients, 2) if patients[i][0] != patients[j][0]],

   [w[k] == Sum(c * (p[i]==k) for i, (_, c) in enumerate(patients)) for k in range(nNurses)],

   # tag(symmetry-breaking)
   [p[z] == z for z in range(nZones)],

   Increasing(w)
)

minimize(
    Sum(w[k] * w[k] for k in range(nNurses))
)
\end{python}\end{boxpy}

This problem involves 2 arrays of variables and 3 types of constraints: \gb{Cardinality}, \gb{Sum} and \gb{Intension}.
A series of 12 instances has been selected for the competition (coming from data generated by P. Schaus).
For generating an \x3 instance (file), you can execute for example:
\begin{command}
python NurseWorkload.py -data=2zones0.txt -parser=NursingWorkload_Parser.py
\end{command}
where `2zones0.txt' is a data file and `NursingWorkload\_ParserZ.py' is a parser (i.e., a Python file allowing us to load data that are not directly given in JSON format).
Note that for saving data in JSON files, you can add the option `-export' (or `-dataexport').

\subsection{RCPSP}

This is Problem \href{https://www.csplib.org/Problems/prob061}{061} on CSPLib, called Resource-Constrained Project Scheduling Problem (RCPSP).
See also \href{https://www.om-db.wi.tum.de/psplib/}{PSPLIB}.

\paragraph{Description {\small (excerpt from CSPLib)}.}
The resource-constrained project scheduling problem is a classical well-known problem in operations research.
  A number of activities are to be scheduled. Each activity has a duration and cannot be interrupted.
  There are a set of precedence relations between pairs of activities which state that the second activity must start after the first has finished.
  There are a set of renewable resources.
  Each resource has a maximum capacity and at any given time slot no more than this amount can be in use.
  Each activity has a demand (possibly zero) on each resource.
  The dummy source and sink activities have zero demand on all resources.
The problem is usually stated as an optimisation problem where the makespan (i.e. the completion time of the sink activity) is minimized.

\paragraph{Data.}
As an illustration of data specifying an instance of this problem, we have:

\begin{json}
{
  "horizon": 158,
  "resourceCapacities": [12, 13, 4, 12],
  "jobs": [
    { "duration": 0, "successors": [1, 2, 3], "requiredQuantities": [0, 0, 0, 0] },
    { "duration": 8, "successors": [5, 10, 14], "requiredQuantities": [4, 0, 0, 0] },
    ...,
    { "duration": 0, "successors": [], "requiredQuantities": [0, 0, 0, 0] }
  ]
}
\end{json}

\paragraph{Model.}
The \p3 model, in a file `Rcpsp.py', used for the competition is: 

\begin{boxpy}\begin{python}
@\imp@
    
jobs, horizon, capacities, _ = data
durations, successors, quantities = zip(*jobs)  # [job.duration for job in jobs]
nJobs = len(jobs)

# s[i] is the starting time of the ith job
s = VarArray(size=nJobs, dom=lambda i: {0} if i == 0 else range(horizon))

satisfy(
   # precedence constraints
   [s[i] + durations[i] <= s[j] for i in range(nJobs) for j in successors[i]],

   # resource constraints
   [
      Cumulative(
         Task(origin=s[i], length=durations[i], height=quantities[i][k]) for i in range(nJobs) if quantities[i][k] > 0
      ) <= capacity for k, capacity in enumerate(capacities)
   ]
)

minimize(
   s[-1]
)
\end{python}\end{boxpy}

This model involves 1 array of variables and 2 types of constraints: \gb{Cumulative} and \gb{Intension}.
A series of 10 instances has been selected for the competition.
For generating an \x3 instance (file), you can execute for example:
\begin{command}
python Rcpsp.py -data=j120-01-01.sm -parser=Rcpsp_Parser.py
\end{command}
where `j120-01-01.sm' is a data file and `Rcpsp\_ParserZ.py' is a parser (i.e., a Python file allowing us to load data that are not directly given in JSON format).
Note that for saving data in JSON files, you can add the option `-export' (or `-dataexport').

\subsection{RLFAP}

When radio communication links are assigned the same or closely related frequencies, there is a potential for interference.
Consider a radio communication network, defined by a set of radio links.
The radio link frequency assignment problem \cite{CGLSW_radio} is to assign, from limited spectral resources, a frequency to each of these links in such a way that all the links may operate together without noticeable interference.
Moreover, the assignment has to comply to certain regulations and physical constraints of the transmitters.
Among all such assignments, one will naturally prefer those which make good use of the available spectrum, trying to save the spectral resources for a later extension of the network.
As in 2018, do note that we consider here the original COP instances.

\paragraph{Description}
The description is rather complex. Hence, we refer the reader to \cite{CGLSW_radio}.

\paragraph{Data.}

As an illustration of data specifying an instance of this problem, we have:
\begin{json}
{
  "domains": {
    "1": [16, 30, 44, 58, 72, 86, 100, 114, 128, 142, 156, 254, 268, ...],
    "2":[30, 58, 86, 114, 142, 268, 296, 324, 352, 380, 414, 442, 470, ...],
    ...
  },
  "vars": [
    { "number": 13, "domain": 1, "value": "null", "mobility": "null" },
    { "number": 14, "domain": 1, "value": "null", "mobility": "null" },
    ...
  ],
  "ctrs":[
    { "x": 13, "y": 14, "equality": true, "limit": 238, "weight": 0 },
    { "x": 13, "y": 16, "equality": false, "limit": 186, "weight": 0 },
    ...
  ],
  "interferenceCosts": [0, 1000, 100, 10, 1],
  "mobilityCosts": [0, 0, 0, 0, 0]
}
\end{json}

\paragraph{Model.}

The \p3 model, in a file `Rlfap.py', used for the competition is: 

\begin{boxpy}\begin{python}
@\imp@

domains, variables, constraints, interferenceCosts, mobilityCosts = data
n = len(variables)

# f[i] is the frequency of the ith radio link
f = VarArray(size=n, dom=lambda i: domains[variables[i].domain])

satisfy(
   # managing pre-assigned frequencies
   [f[i] == v for i, (_, v, mob) in enumerate(variables)
                if v and not (variant("max") and mob)],

   # hard constraints on radio-links
   [expr(op, abs(f[i] - f[j]), k) for (i, j, op, k, wgt) in constraints
      if not (variant("max") and wgt)]
)

   
if variant("span"):
   minimize(
      # minimizing the largest frequency
      Maximum(f)
   )

elif variant("card"):
   minimize(
      # minimizing the number of used frequencies
      NValues(f)
   )

elif variant("max"):
   minimize(
      # minimizing the sum of violation costs
      Sum(ift(f[i] == v, 0, mobilityCosts[mob])
           for i, (_, v, mob) in enumerate(variables) if v and mob)
      + Sum(ift(expr(op, abs(f[i] - f[j]), k), 0, interferenceCosts[wgt])
           for (i, j, op, k, wgt) in constraints if wgt)
   )
\end{python}\end{boxpy}

The model involves 1 array of variables,  many constraints \gb{Intension} and an objective that varies according to the chosen variant.
Note that \texttt{expr} allows us to build an expression (constraint) with an operator given as first parameter (possibly, a string).
The complete series of 25 instances, 11 CELAR (scen) and 14 GRAPH, has been selected for the competition.
For generating an \x3 instance (file), you can execute for example:
\begin{command}
python Rlfap.py -data=Rlfap-span-graph-03.json -variant=span
python Rlfap.py -data=Rlfap-card-graph-01.json -variant=card 
python Rlfap.py -data=Rlfap-max-graph-05.json -variant=max
\end{command}

\subsection{Spot5}

\paragraph{Description.}
A model for the following problem has been originally written by Simon de Givry for the 2014 and 2015 Minizinc competitions.
The problem is roughly described (by Simon) as follows:
\begin{itemize}
\item given a set $S$ of photographs which can be taken the next day from at least one of the three instruments,  w.r.t. the satellite trajectory;
\item given, for each photograph, a  weight  expressing its importance;
\item given a set of imperative constraints:  non overlapping and minimal transition time  between two successive  photographs on the  sameinstrument, limitation on the instantaneous data flow through the satellite telemetry and on the recording capacity on board;
\end{itemize}
find  an admissible subset $S'$ of $S$ (imperative  constraints met) which maximizes the sum of the weights of the photographs in S'.
See also \cite{spot5}.

\paragraph{Data.}
As an illustration, the structure of the data specifying an instance of this problem is:

\begin{json}
{
  "domains": ...,
  "costs": ...,
  "c2s": ...,
  "c3s": ...
}
\end{json}

\paragraph{Model.}
The \p3 model, in a file `Spot5.py', used for the competition is: 

\begin{boxpy}\begin{python}
@\imp@

domains, costs, c2s, c3s = data
n = len(domains)

# x[i] is the value for the ith variable
x = VarArray(size=n, dom=lambda i: domains[i])

satisfy(
   # binary constraints
   [(x[i], x[j]) in [tuple(t[i * 2:i * 2 + 2]) for i in range(len(t) // 2)]
      for i, j, t in c2s],

   # ternary constraints
   [(x[i], x[j], x[k]) in [tuple(t[i * 3:i * 3 + 3]) for i in range(len(t) // 3)]
      for i, j, k, t in c3s]
)
   
minimize(
   Sum(costs[i] * (x[i] == 0) for i in range(n))
)
\end{python}\end{boxpy}

This problem involves 1 array of variables and 1 type of constraints: \gb{Extension}.
A series of 10 instances has been selected for the competition (coming from data generated by S. de Givry for Minizinc competitions).
For generating an \x3 instance (file), you can execute for example:
\begin{command}
python Spot5.py -data=0412.json 
\end{command}

\subsection{TAL}

TAL is a problem of natural language processing.
The series of instances, used for the 2022 competition comes from the 2018 competition. The model has not been converted in \p3.


\subsection{Triangular}

\paragraph{Description.}
This problem, taken from Daily Telegraph and Sunday Times, is to find, for an equilateral triangular grid of size $n$ (length of a side),
the maximum number of nodes that can be selected without having all selected corners of any equilateral triangle of any size or orientation.
It was also used in Minizinc Competitions (2015 and 2019).

\paragraph{Data.}
Only one integer is required to specify a specific instance. Values of $n$ used for the instances in the competition are:
\begin{quote}
10, 15, 20, 22, 25, 28, 30, 32, 35, 38
\end{quote}

\paragraph{Model.}
The \p3 model, in a file `Triangular.py', used for the competition is: 

\begin{boxpy}\begin{python}
@\imp@

n = data

# x[i,j] is 1 iff the jth node in the ith row is selected
x = VarArray(size=[n, n], dom=lambda i, j: {0, 1} if i >= j else None)

satisfy(
   # avoiding the three corners of any equilateral triangle to be selected
   Sum(
      x[i + m][j],
      x[i + k][j + m],
      x[i + k - m][j + k - m]
   ) <= 2 for i in range(n) for j in range(i + 1) for k in range(1, n - i) for m in range(k)
)

maximize(
   Sum(x)
)
\end{python}\end{boxpy}

This problem involves 1 array of variables and 1 type of constraints: \gb{Sum}.
A series of 10 instances has been selected for the competition.
For generating an \x3 instance (file), you can execute for example:
\begin{command}
python Triangular.py -data=25
\end{command}

\subsection{Warehouse}

This is Problem \href{https://www.csplib.org/Problems/prob034}{034} on CSPLib, called Warehouse Location Problem.

\paragraph{Description {\small (from CSPLib)}.}

In the Warehouse Location problem (WLP), a company considers opening warehouses at some candidate locations in order to supply its existing stores.
Each possible warehouse has the same maintenance cost, and a capacity designating the maximum number of stores that it can supply.
Each store must be supplied by exactly one open warehouse.
The supply cost to a store depends on the warehouse.
The objective is to determine which warehouses to open, and which of these warehouses should supply the various stores, such that the sum of the maintenance and supply costs is minimized.

\paragraph{Data.}

As an illustration of data specifying an instance of this problem, we have:
\begin{json}
{
  "fixedCost": 30,
  "warehouseCapacities": [1,4,2,1,3],
  "storeSupplyCosts": [
    [20,24,11,25,30],[28,27,82,83,74],[74,97,71,96,70],[2,55,73,69,61],
    [46,96,59,83,4],[42,22,29,67,59],[1,5,73,59,56],[10,73,13,43,96],
    [93,35,63,85,46],[47,65,55,71,95]
  ]
}
\end{json}

\paragraph{Model.}
The \p3 model, in a file `Warehouse.py', used for the competition is: 

\begin{boxpy}\begin{python}
@\imp@

cost, capacities, costs = data  # cost is the fixed cost when opening a warehouse
nWarehouses, nStores = len(capacities), len(costs)

# w[i] is the warehouse supplying the ith store
w = VarArray(size=nStores, dom=range(nWarehouses))

# o[j] is 1 if the jth warehouse is open
o = VarArray(size=nWarehouses, dom={0, 1})

# c[i] is the cost of supplying the ith store
c = VarArray(size=nStores, dom=lambda i: costs[i])

satisfy(
   # capacities of warehouses must not be exceeded
   Count(w, value=j) <= capacities[j] for j in range(nWarehouses)

   # the warehouse supplier of the ith store must be open
   [o[w[i]] == 1 for i in range(nStores)],
   
   # computing the cost of supplying the ith store
   [costs[i][w[i]] == c[i] for i in range(nStores)]
)

minimize(
   # minimizing the overall cost
   Sum(c) + Sum(o) * cost
)
\end{python}\end{boxpy}

This model involves 3 arrays of variables and 2 type of constraints: \gb{Count} and \gb{Element}.
A series of 9 instances has been selected for the competition.
For generating an \x3 instance (file), you can execute for example:
\begin{command}
python Warehouse.py -parser=Warehouse_Random.py 40 80 100 10 1000 0
\end{command}
where `Warehouse\_Random.py' is a generator of instances, using specified values.
Note that for saving data in JSON files, you can add the option `-export' (or `-dataexport').

\subsection{War or Peace}

\paragraph{Description.}
Problem based on information from \href{http://www.hakank.org/}{hakank.org}.
There are $n$ countries.
Each pair of two countries is either at war or has a peace treaty.
Each pair of two countries that has a common enemy has a peace treaty.
What is the minimum number of peace treaties?

\paragraph{Data.}
Only one integer is required to specify a specific instance. Values of $n$ used for the instances in the competition are:
\begin{itemize}
\item 8, 9, 10, 12, 14 for main variant
\item 8, 9, 10, 12, 14 for variant 'or'
\end{itemize}

\paragraph{Model.}
The \p3 model, in a file `WarOrPeace.py', used for the competition is: 

\begin{boxpy}\begin{python}
@\imp@

n = data 
WAR, PEACE = 0, 1

# x[i][j] is 1 iff countries i and j have a peace treaty
x = VarArray(size=[n, n], dom=lambda i, j: {WAR, PEACE} if i < j else None)

if not variant():
   satisfy(
      If(
         x[i][j] != PEACE,
         Then=NotExist(both(x[min(i, k)][max(i, k)] == WAR, x[min(j, k)][max(j, k)] == WAR) for k in range(n) if different_values(i, j, k))
      ) for i, j in combinations(n, 2)
   )

elif variant("or"):
   satisfy(
      If(
         x[i][j] != PEACE,
         Then=[
            x[i][j] == WAR,
            conjunction(either(x[k][i] == PEACE, x[k][j] == PEACE) for k in range(i))]
      ) for i, j in combinations(range(1, n), 2)
   )

minimize(
   # minimizing the number of peace treaties
   Sum(x)
)
\end{python}\end{boxpy}

This problem involves 1 array of variables and complex forms of constraints.
A series of 10 instances has been selected for the competition.
For generating an \x3 instance (file), you can execute for example:
\begin{command}
python WarOrPeace.py -data=10
python WarOrPeace.py -data=10 -variant=or
\end{command}

\chapter{Solvers}

In this chapter, we introduce the solvers and teams having participated to the \x3 Competition 2022.

\begin{itemize}
\item ACE (Christophe Lecoutre)
\item ACE ABD (extension of ACE by Thibault Falque and Hughes Wattez)
\item BTD, miniBTD (Mohamed Sami Cherif, Djamal Habet, Philippe J\'egou, H\'el\`ene Kanso, Cyril Terrioux)
\item Choco (Charles Prud'homme and Jean-Guillaume Fages)
\item CoSoCo (Gilles Audemard)
\item Exchequer (Martin Mariusz Lester)
\item Fun-sCOP (Takehide Soh, Daniel Le Berre, Hidetomo Nabeshima, Mutsunori Banbara, Naoyuki Tamura)
\item Glasgow (Ciaran McCreesh)
\item MiniCPBP (Gilles Pesant and Auguste Burlats)
\item Mistral (Emmanuel Hebrard and Mohamed Siala)
\item Nacre (Ga\"el Glorian)
\item Picat (Neng-Fa Zhou)
\item RBO, miniRBO (Mohamed Sami Cherif, Djamal Habet, Cyril Terrioux)
\item Sat4j-CSP-PB (extension of Sat4j by Thibault Falque and Romain Wallon)
\item toulbar2 (David Allouche et al.) 
\end{itemize}

\addcontentsline{toc}{section}{\numberline{}ACE}
\includepdf[pages=-,pagecommand={\thispagestyle{plain}}]{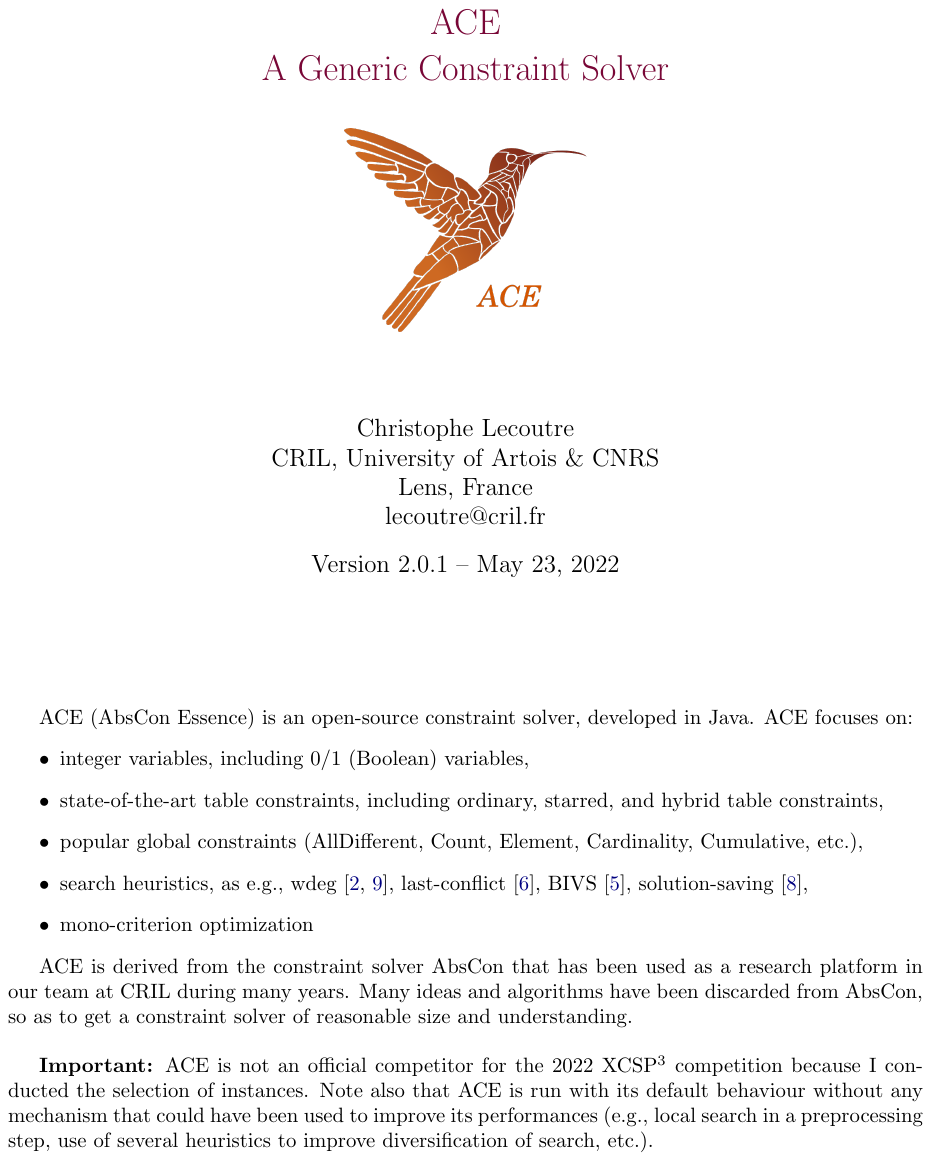}
\addcontentsline{toc}{section}{\numberline{}ACE ABD}
\includepdf[pages=-,pagecommand={\thispagestyle{plain}}]{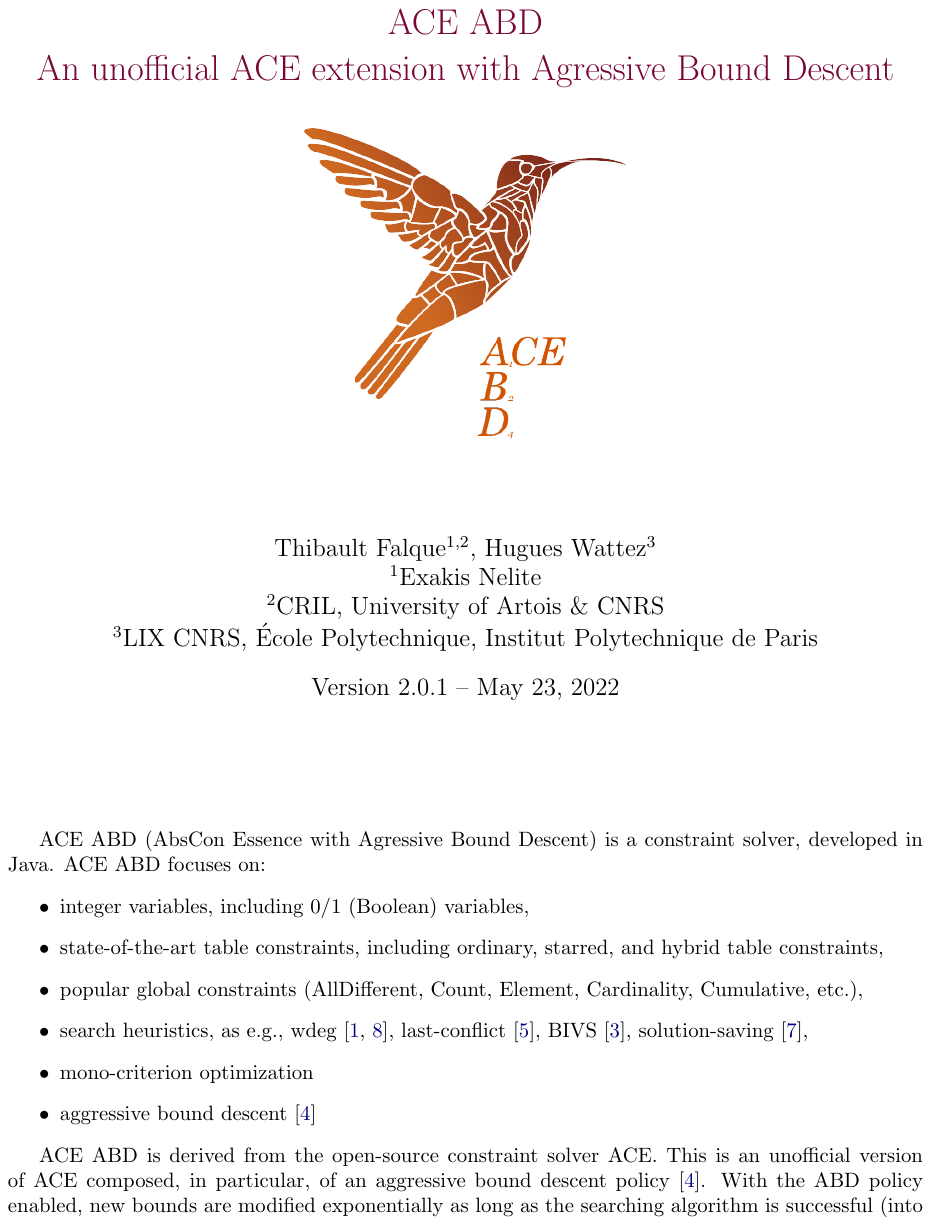}
\addcontentsline{toc}{section}{\numberline{}BTD}
\includepdf[pages=-,pagecommand={\thispagestyle{plain}}]{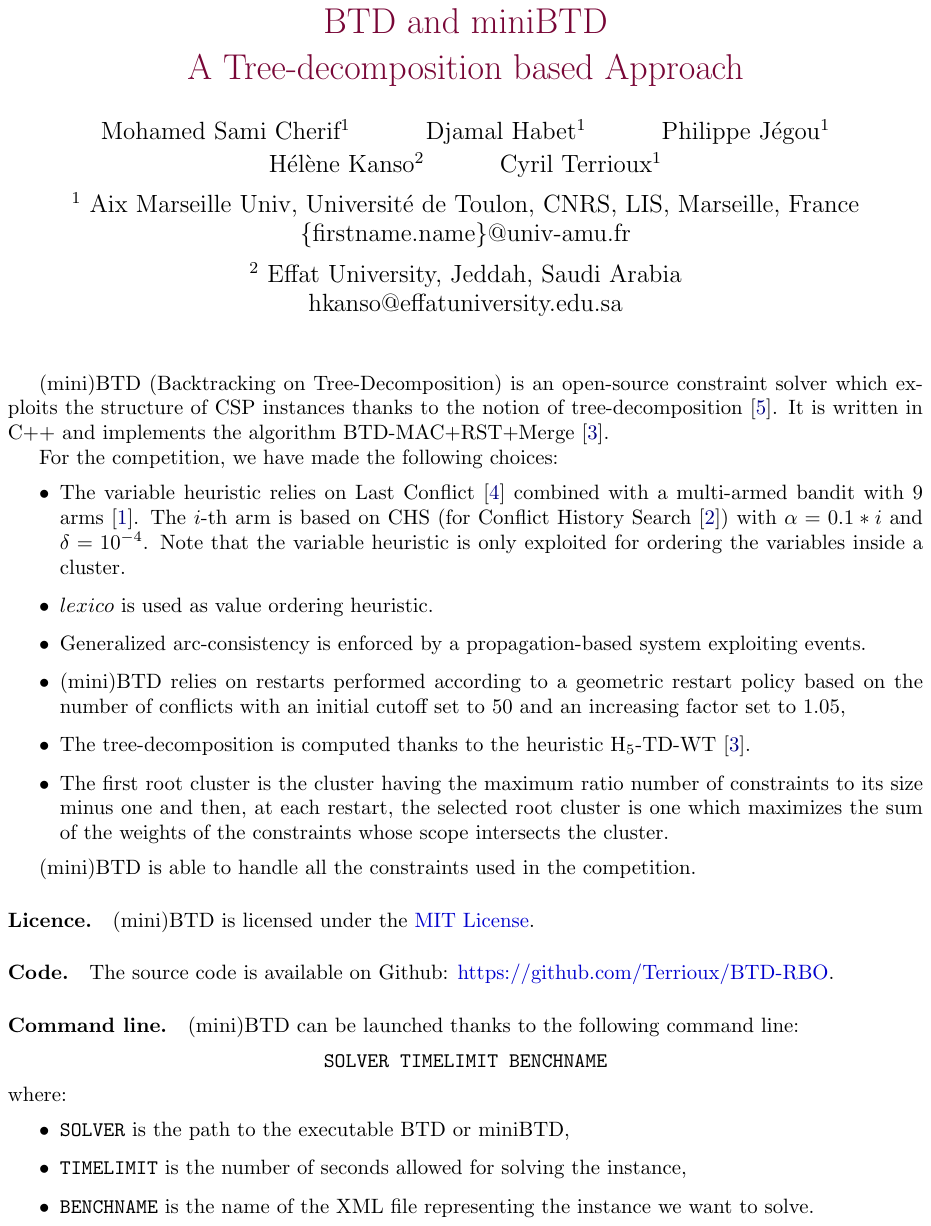}
\addcontentsline{toc}{section}{\numberline{}Choco}
\includepdf[pages=-,pagecommand={\thispagestyle{plain}}]{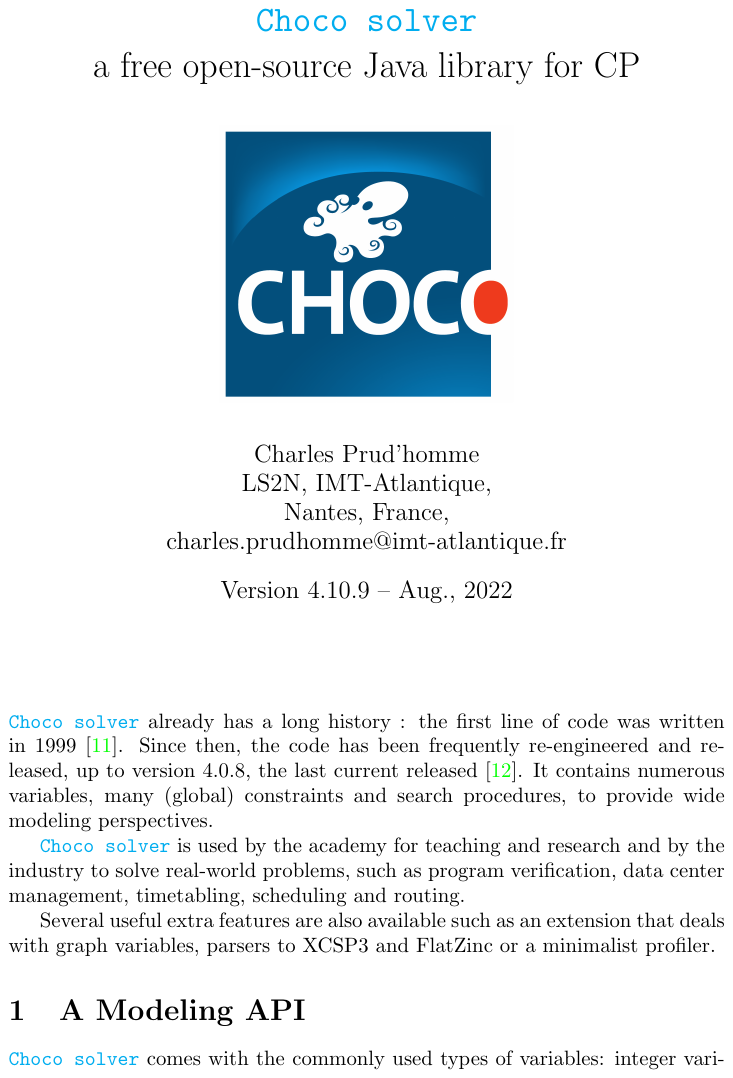}
\addcontentsline{toc}{section}{\numberline{}CoSoCo}
\includepdf[pages=-,pagecommand={\thispagestyle{plain}}]{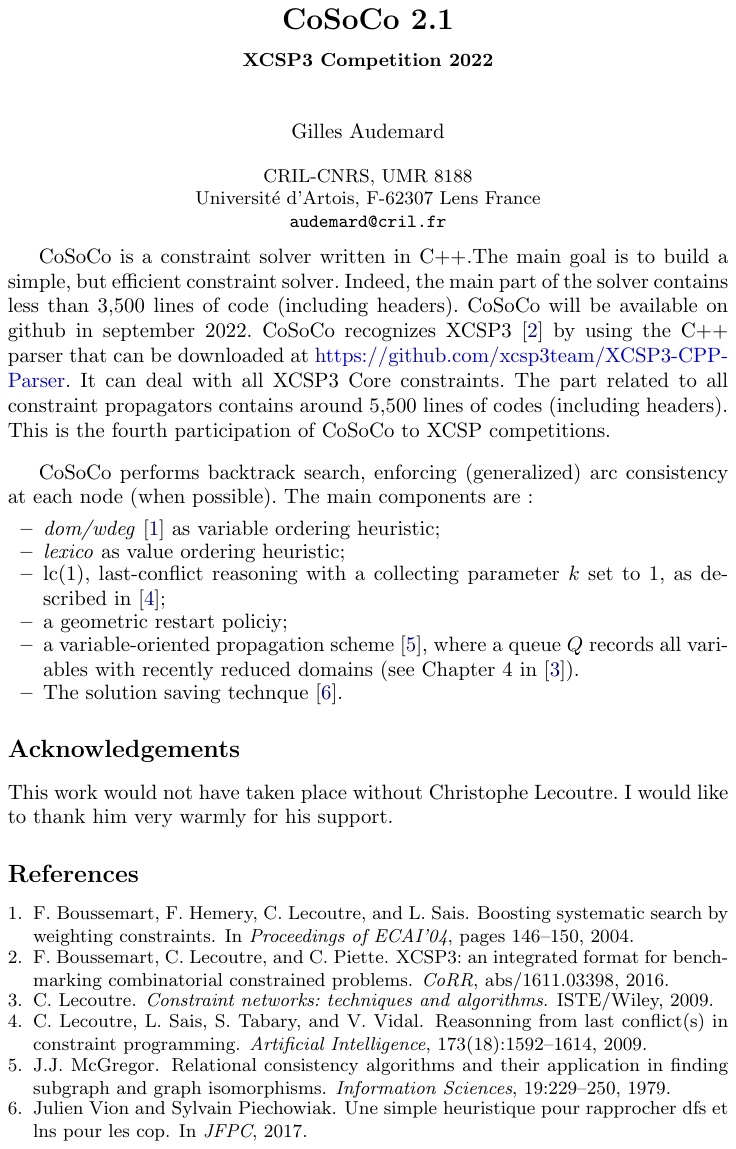}
\addcontentsline{toc}{section}{\numberline{}Exchequer}
\includepdf[pages=-,pagecommand={\thispagestyle{plain}}]{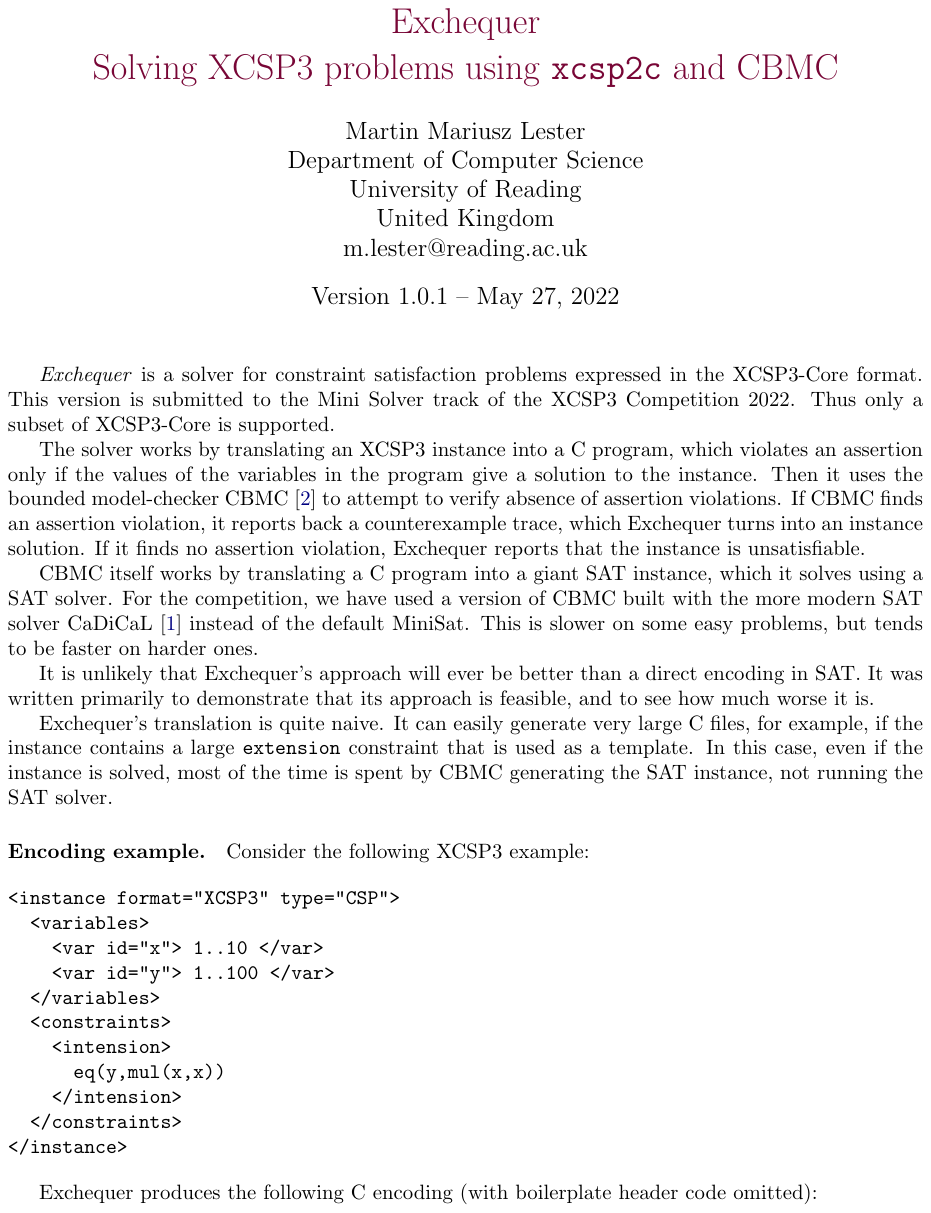}
\addcontentsline{toc}{section}{\numberline{}Fun-sCOP}
\includepdf[pages=-,pagecommand={\thispagestyle{plain}}]{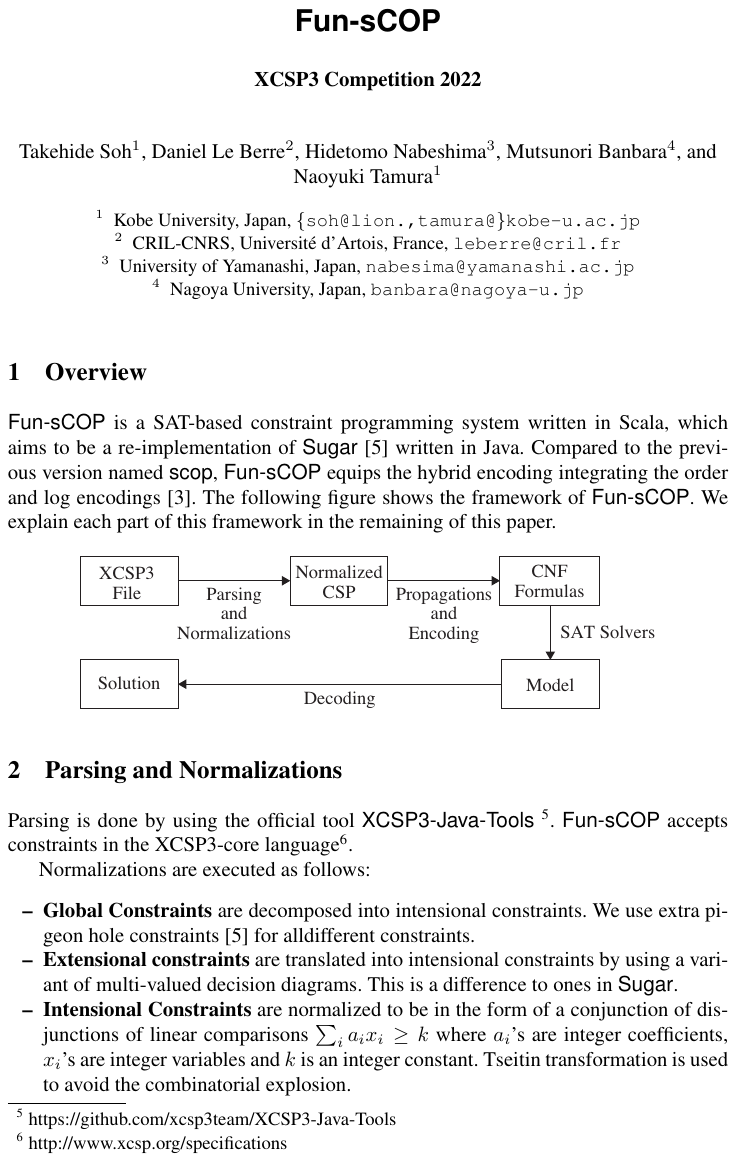}
\addcontentsline{toc}{section}{\numberline{}Glasgow}
\includepdf[pages=-,pagecommand={\thispagestyle{plain}}]{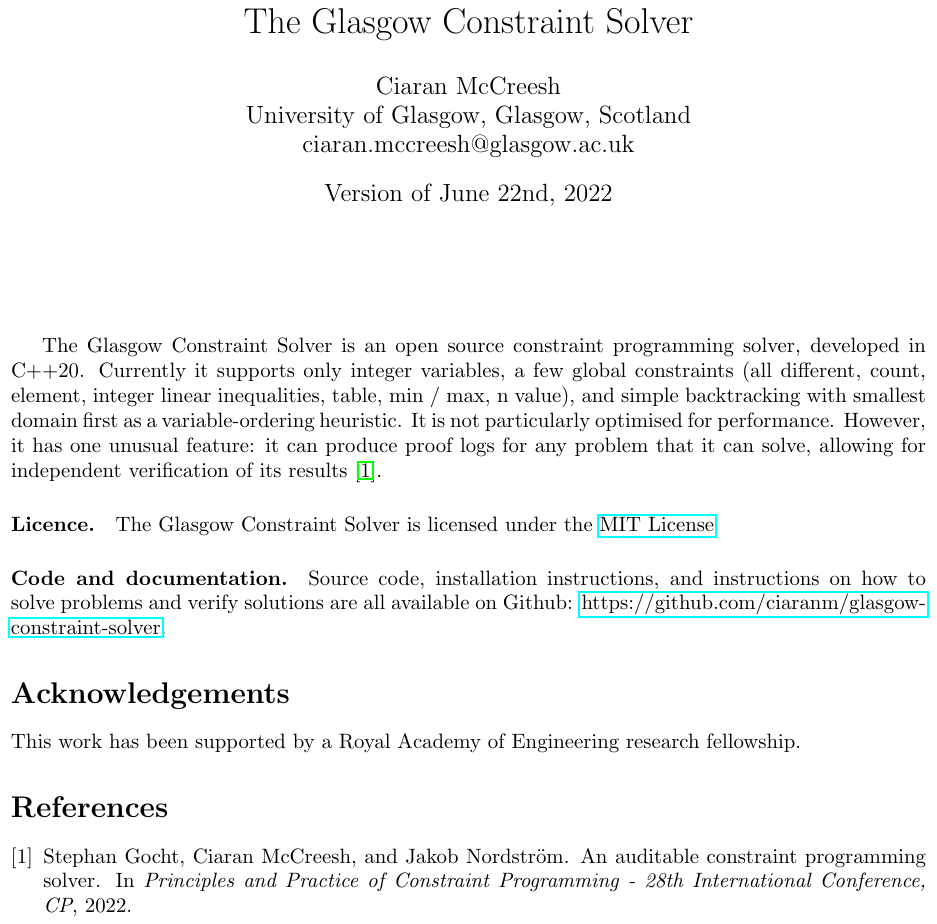}
\addcontentsline{toc}{section}{\numberline{}MiniCPBP}
\includepdf[pages=-,pagecommand={\thispagestyle{plain}}]{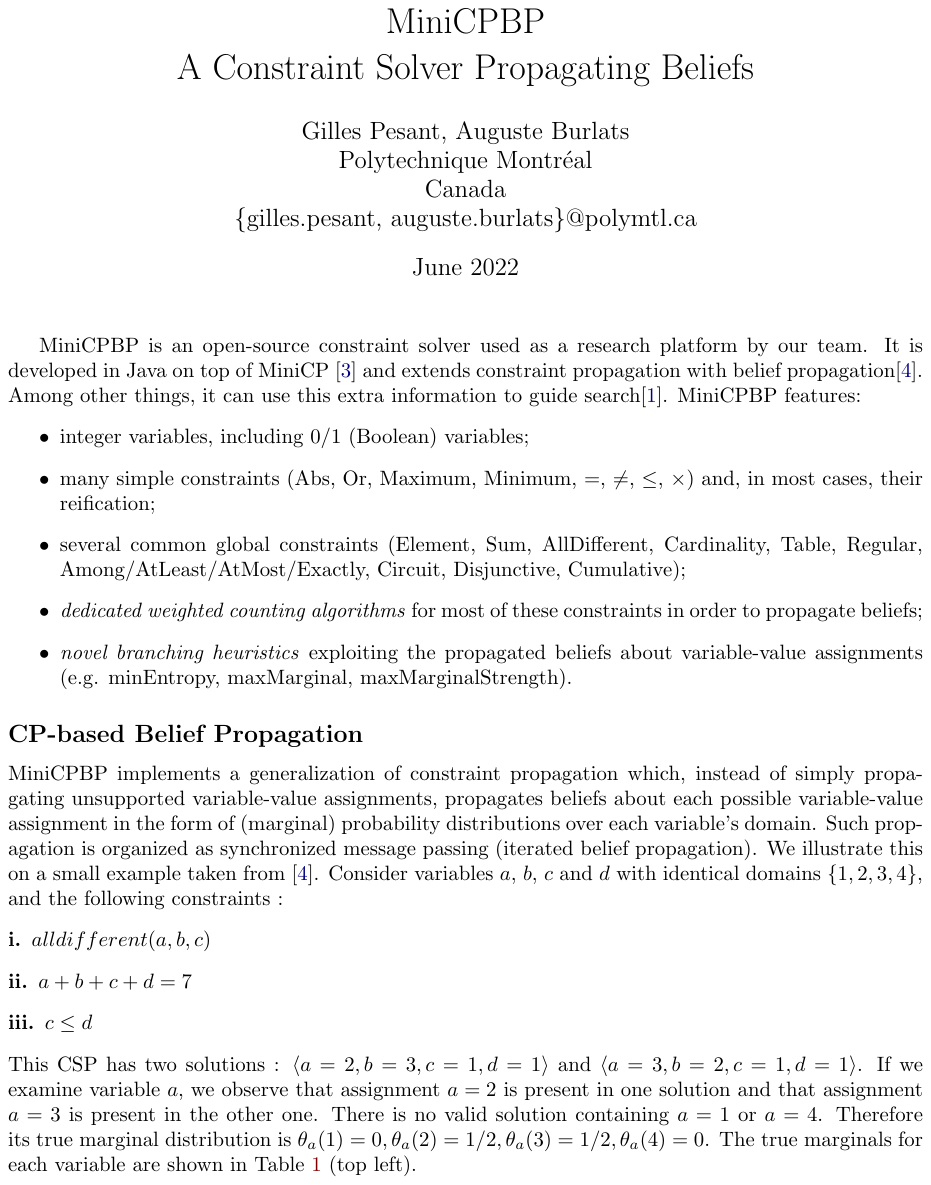}
\addcontentsline{toc}{section}{\numberline{}Mistral}
\includepdf[pages=-,pagecommand={\thispagestyle{plain}}]{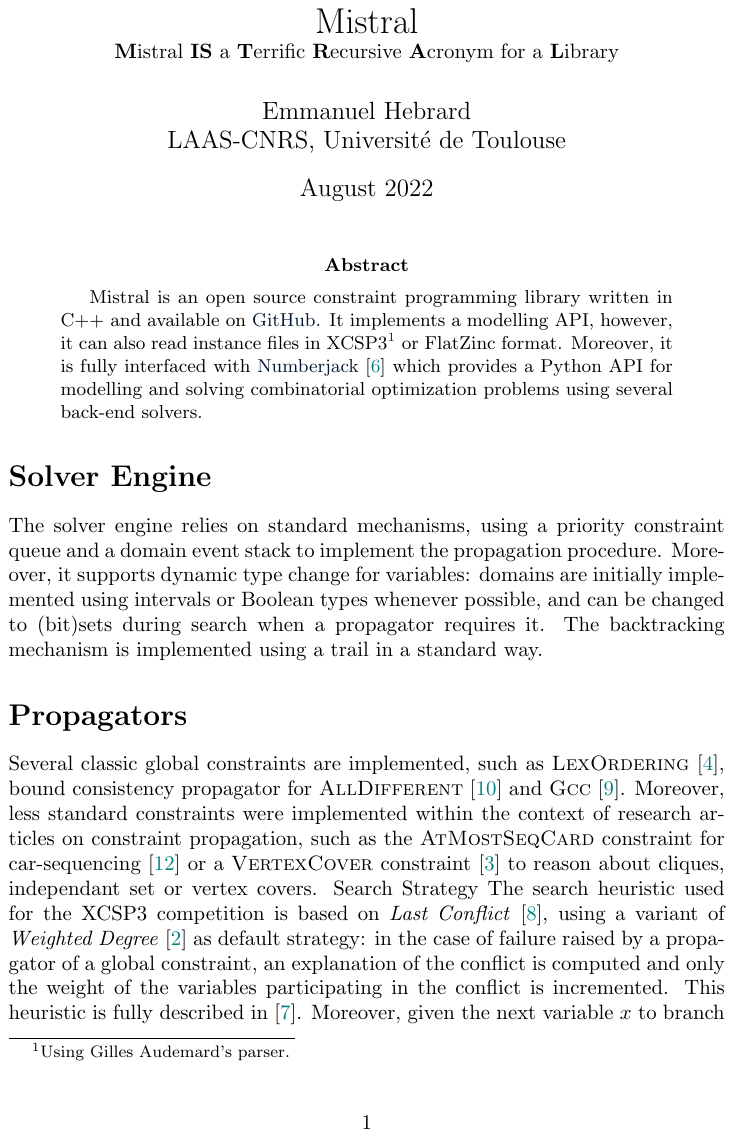}
\addcontentsline{toc}{section}{\numberline{}NACRE}
\includepdf[pages=-,pagecommand={\thispagestyle{plain}}]{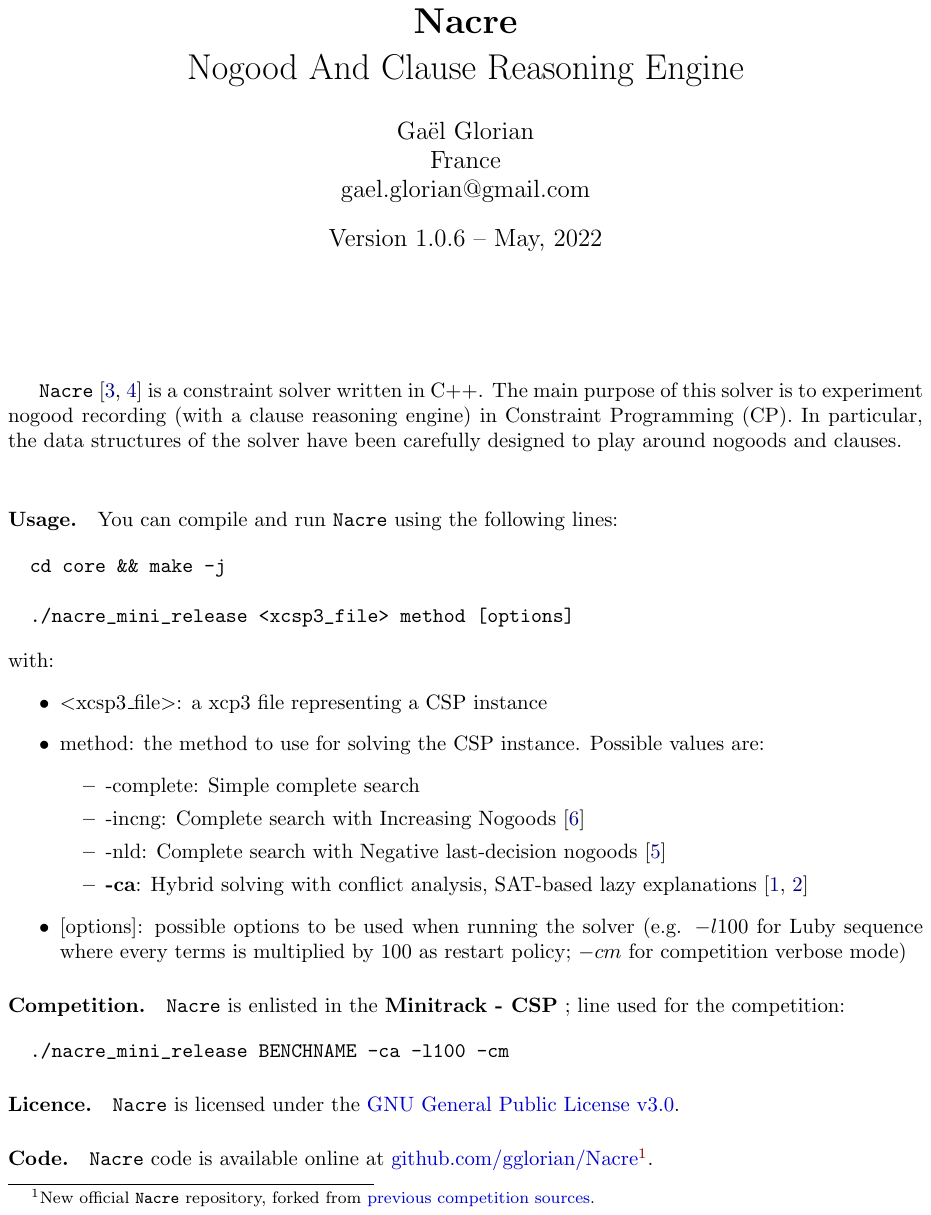}
\addcontentsline{toc}{section}{\numberline{}Picat}
\includepdf[pages=-,pagecommand={\thispagestyle{plain}}]{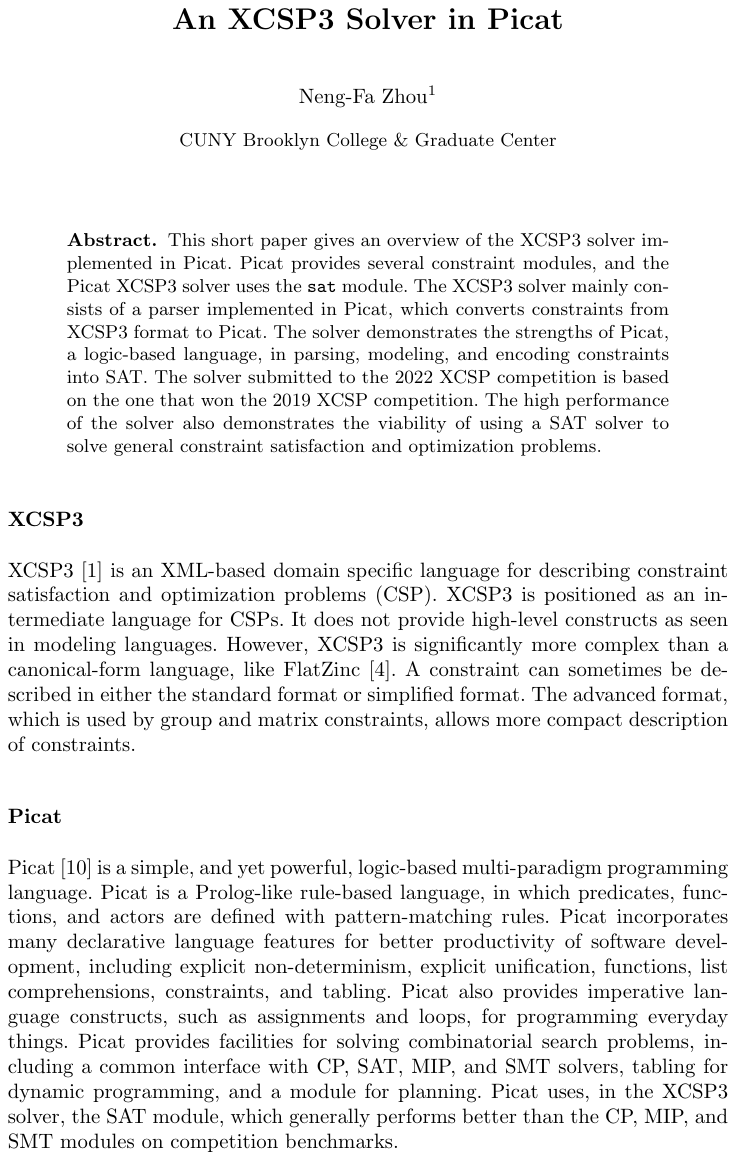}
\addcontentsline{toc}{section}{\numberline{}RBO}
\includepdf[pages=-,pagecommand={\thispagestyle{plain}}]{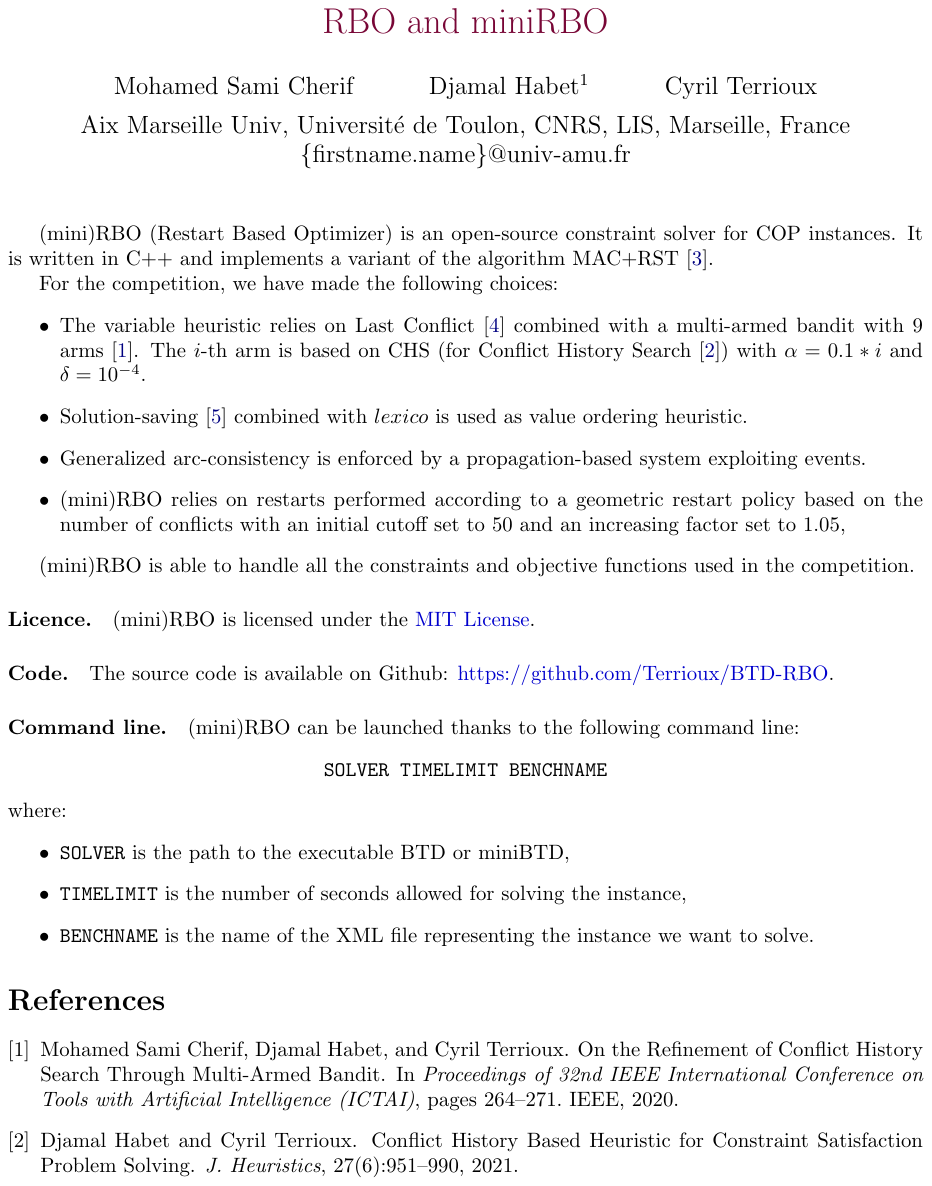}
\addcontentsline{toc}{section}{\numberline{}Sat4j-CSP-PBj}
\includepdf[pages=-,pagecommand={\thispagestyle{plain}}]{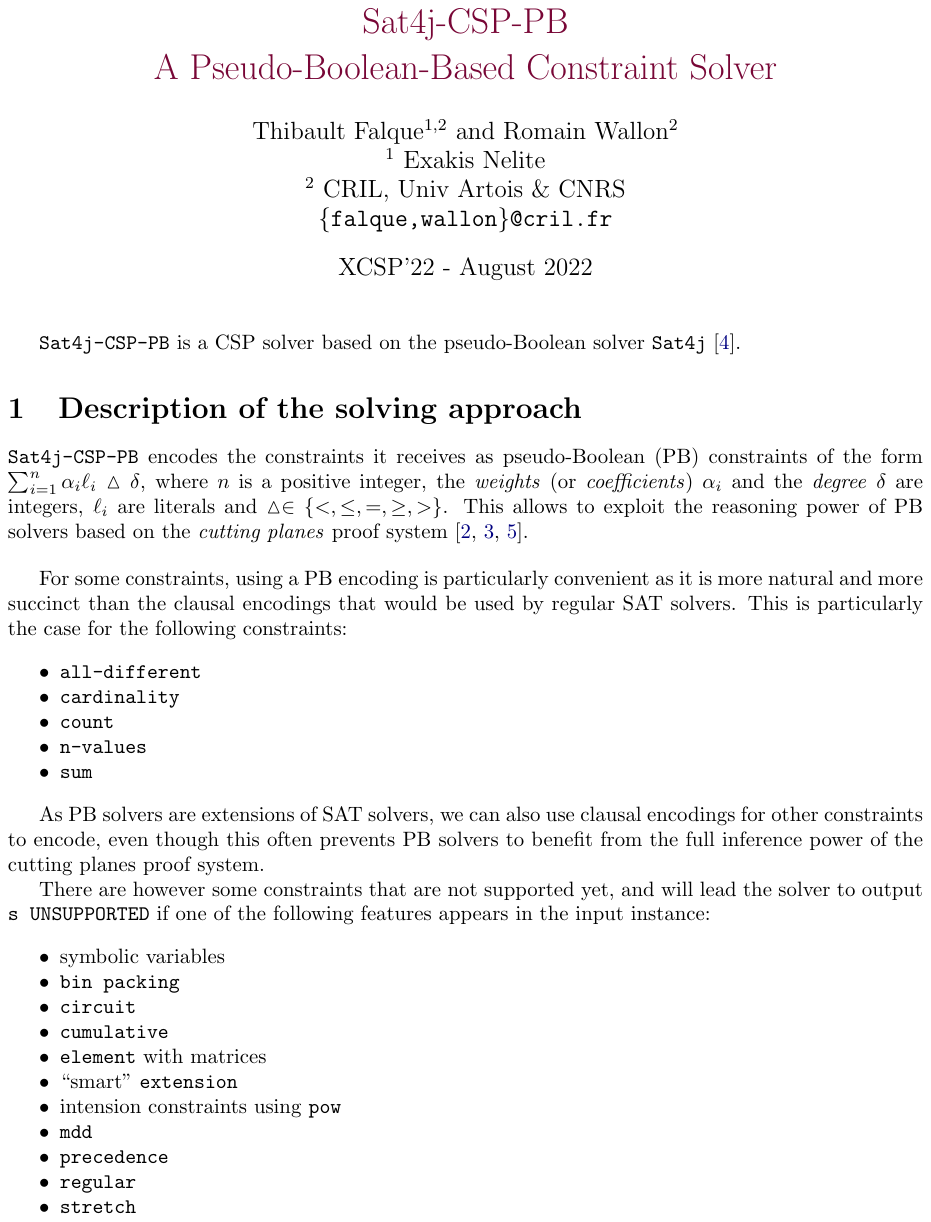}
\addcontentsline{toc}{section}{\numberline{}toulbar2}
\includepdf[pages=-,pagecommand={\thispagestyle{plain}}]{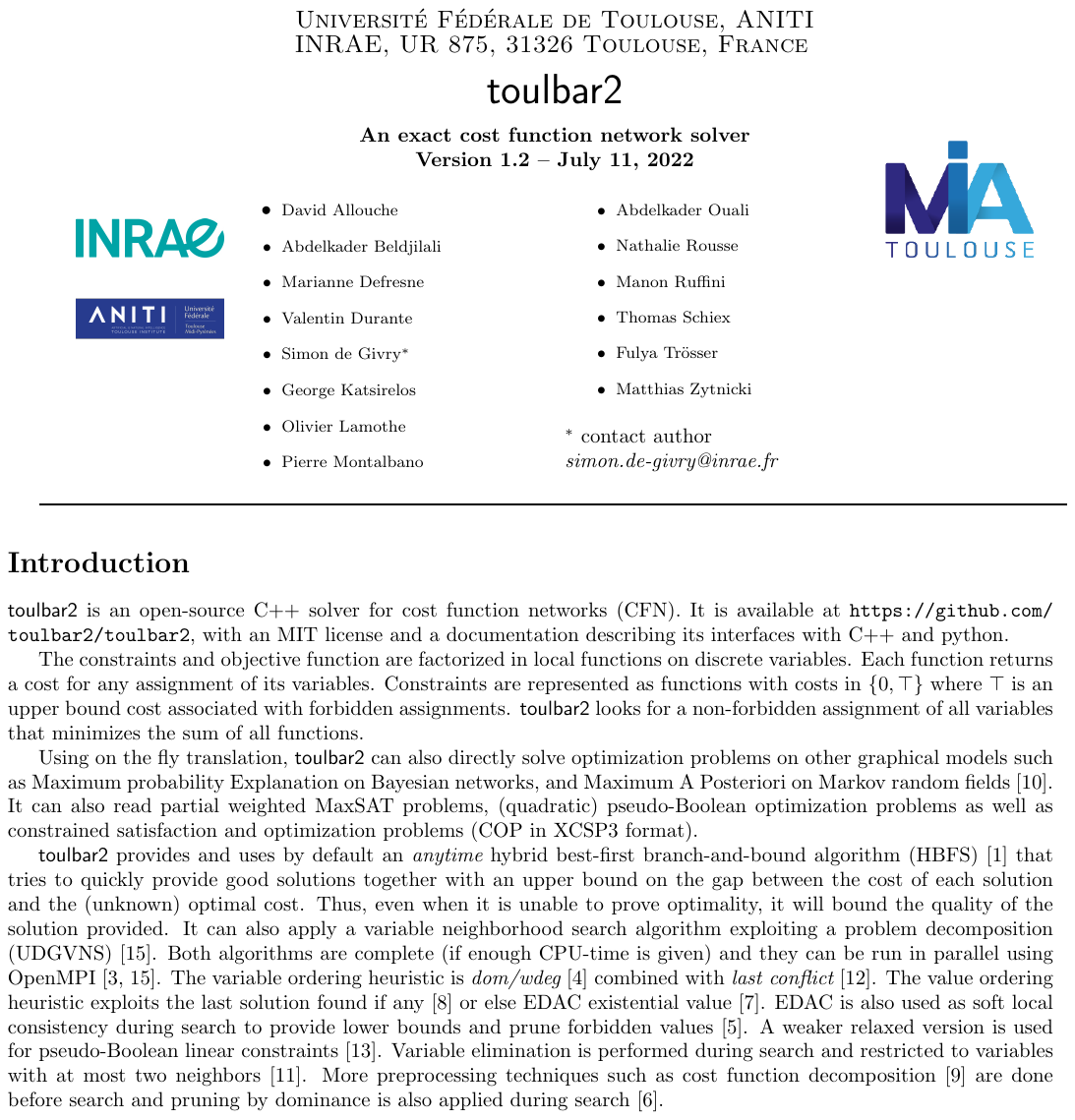}


\chapter{Results}

In this chapter, rankings for the various tracks of the \x3 Competition 2022 are given.
Importantly, remember that you can find all detailed results, including all traces of solvers at \href{http://www.cril.univ-artois.fr/XCSP22/}{http://www.cril.univ-artois.fr/XCSP22/}.

\section{Context}

\bigskip
Remember that the tracks of the competition are given by Table \ref{tab:anysolver} and Table \ref{tab:minisolver}.

\begin{table}[h!]
\begin{center}
\begin{tabular}{cccc} 
\toprule
\textcolor{dred}{\bf Problem} &  \textcolor{dred}{\bf Goal} &  \textcolor{dred}{\bf Exploration} &  \textcolor{dred}{\bf Timeout} \\
\midrule
CSP  & one solution & sequential & 40 minutes \\
COP  & best solution & sequential & 40 minutes \\
fast COP  & best solution & sequential & 4 minutes \\
// COP  & best solution & parallel & 40 minutes \\
\bottomrule
\end{tabular}
\end{center}
\caption{Standard Tracks. \label{tab:anysolver}}
\end{table}

\begin{table}[h!]
\begin{center}
\begin{tabular}{cccc} 
\toprule
\textcolor{dred}{\bf Problem} &  \textcolor{dred}{\bf Goal} &  \textcolor{dred}{\bf Exploration} &  \textcolor{dred}{\bf Timeout} \\
\midrule
MiniCSP  & one solution & sequential & 40 minutes \\
MiniCOP  & best solution & sequential & 40 minutes \\
\bottomrule
\end{tabular}
\end{center}
\caption{Mini-Solver Tracks. \label{tab:minisolver}}
\end{table}

\noindent Also, note that:

\begin{itemize}
\item the cluster was provided by CRIL and is composed of nodes with two quad-cores (Intel Xeon CPU E5-2637 v4 @ 3.50GHz, each equipped with 64 GiB RAM).
\item Hyperthreading was disabled.
\item Each solver was allocated a CPU and 64 GiB of RAM, independently from the tracks.
\item Timeouts were set accordingly to the tracks through the tool \texttt{runsolver}:
  \begin{itemize}
    \item sequential solvers in the fast COP track were allocated 4 min of CPU time and 12~min of Wall Clock time,
    \item other sequential solvers were allocated 40 min of CPU time and 120 min of Wall Clock time,
    \item parallel solvers were allocated 160 min of CPU time and 120 min 
      of Wall Clock time.
  \end{itemize}
\item The selection of instances for the Standard tracks was composed of 200 CSP instances and 250~COP instances.
\item The selection of instances for the Mini-solver tracks was composed of 150 CSP instances and 158~COP instances.
\end{itemize}

\paragraph{About Scoring.}
The number of points won by a solver $S$ is decided as follows:
\begin{itemize}
\item for CSP, this is the number of times $S$ is able to solve an instance, i.e., to decide the satisfiability of an instance (either exhibiting a solution, or indicating that the instance is unsatisfiable)
\item for COP, this is, roughly speaking, the number of times $S$ gives the best known result, compared to its competitors.
More specifically, for each instance $I$:
\begin{itemize}
\item if $I$ is unsatisfiable, 1 point is won by $S$ if $S$ indicates that the instance $I$ is unsatisfiable, 0 otherwise,
\item if $S$ provides a solution whose bound is less good than another one (found by another competiting solver), 0 point is won by $S$,
\item if $S$ provides an optimal solution, while indicating that it is indeed the optimality, 1 point is won by $S$,
\item if $S$ provides (a solution with) the best found bound among all competitors, this being possibly shared by some other solver(s), while indicating no information about optimality: 1 point is won by $S$ if no other solver proved that this bound was optimal, $0.5$ otherwise.
\end{itemize}
\end{itemize}


\paragraph{Off-competition Solvers.} Some solvers were run while not officially entering the competition: we call them {\em off-competition} solvers.
\ace is one of them because its author (C. Lecoutre) conducted the selection of instances, which is a very strong bias (ACE\_ABD was also considered as being off-competition of the main tracks to avoid any suspected collusion).
Also, when two variants (by the same competiting team) of a same solver compete in a same track, only the best one is ranked (and the second one considered as being off-competition).
This is why Fun-sCOP-glue was considered as off-competition in the CSP track.

\section{Rankings}







\bigskip
Here are the rankings\footnote{The images of medals come from \href{https://freesvg.org/gold-medal-juhele-final}{freesvg.org}} for the 6 tracks.
\bigskip

\begin{minipage}{0.4\textwidth}
\begin{center}
  \begin{tabular}{|lcp{0.0cm}p{2.1cm}|}
    \hline
     \vspace{-0.2cm} & & &  \\
    \multirow{5}{2.1cm}{{\bf {\large ~ CSP}}} & \includegraphics[scale=0.15]{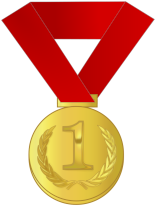} & & \vspace{-0.6cm} {\large Picat} \\
    & & & \\
    & \includegraphics[scale=0.15]{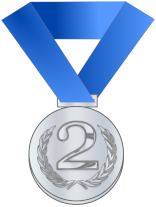} & & \vspace{-0.6cm} {\large Fun-sCOP} \\
    & & & \\
    & \includegraphics[scale=0.15]{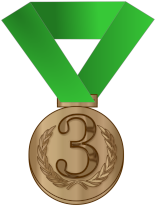} & & \vspace{-0.6cm} {\large Choco} \\
    \hline 
  \end{tabular}
\end{center}
\end{minipage} \hspace{1.4cm}
\begin{minipage}{0.4\textwidth}
\begin{center}
  \begin{tabular}{|lcp{0.0cm}p{2.1cm}|}
    \hline
    \vspace{-0.2cm} & & &  \\
    \multirow{5}{2.1cm}{{\bf {\large ~ COP}}} & \includegraphics[scale=0.15]{gold.png} & & \vspace{-0.6cm}  {\large Picat} \\
    & & & \\
    & \includegraphics[scale=0.15]{silver.png} & & \vspace{-0.6cm} {\large CoSoCo} \\
    & & & \\
    & \includegraphics[scale=0.15]{bronze.png} & & \vspace{-0.6cm} {\large Mistral} \\
    \hline
  \end{tabular}
\end{center}
\end{minipage}

\bigskip

\begin{minipage}{0.4\textwidth}  
\begin{center}
  \begin{tabular}{|lcp{0.0cm}p{2.1cm}|}
    \hline
    \vspace{-0.2cm} & & &  \\
    \multirow{5}{2.1cm}{{\bf {\large Fast COP}}} & \includegraphics[scale=0.15]{gold.png} & & \vspace{-0.6cm} {\large CoSoCo} \\
    & & & \\
    & \includegraphics[scale=0.15]{silver.png} & & \vspace{-0.6cm} {\large Picat} \\
    & & & \\
    & \includegraphics[scale=0.15]{bronze.png} & & \vspace{-0.6cm} {\large Mistral} \\
    \hline
  \end{tabular}
\end{center}
\end{minipage} \hspace{1.4cm}
\begin{minipage}{0.4\textwidth}
\begin{center}
  \begin{tabular}{|lcp{0.0cm}p{2.1cm}|}
    \hline
    \vspace{-0.2cm} & & &  \\
    \multirow{5}{2.1cm}{{\bf {\large // COP}}} & \includegraphics[scale=0.15]{gold.png} & & \vspace{-0.6cm} {\large Choco} \\
    & & & \\
    & \includegraphics[scale=0.15]{silver.png} & & \vspace{-0.6cm} {\large Toulbar2} \\
    & & & \\
    & \includegraphics[scale=0.15]{bronze.png} & & \vspace{-0.6cm} {\large ~ --}  \\
    \hline
  \end{tabular}
\end{center}
\end{minipage}

\bigskip

\begin{minipage}{0.4\textwidth}  
\begin{center}
  \begin{tabular}{|lcp{0.0cm}p{2.1cm}|}
    \hline
    \vspace{-0.2cm} & & &  \\
    \multirow{5}{2.1cm}{{\bf {\large Mini CSP}}} &\includegraphics[scale=0.15]{gold.png} & & \vspace{-0.6cm} {\large Exchequer} \\
    & & & \\
    & \includegraphics[scale=0.15]{silver.png} & & \vspace{-0.6cm} {\large miniBTD} \\
    & & & \\
    & \includegraphics[scale=0.15]{bronze.png} & & \vspace{-0.6cm} {\large Sat4j-CSP} \\
    \hline
  \end{tabular}
\end{center}
\end{minipage} \hspace{1.4cm}
\begin{minipage}{0.4\textwidth}
\begin{center}
  \begin{tabular}{|lcp{0.0cm}p{2.1cm}|}
    \hline
    \vspace{-0.2cm} & & &  \\
    \multirow{5}{2.15cm}{{\bf {\large Mini COP}}} & \includegraphics[scale=0.15]{gold.png} & & \vspace{-0.6cm} {\large Mistral} \\
    & & & \\
    & \includegraphics[scale=0.15]{silver.png} & & \vspace{-0.6cm} {\large Toulbar2} \\
    & & & \\
    & \includegraphics[scale=0.15]{bronze.png} & & \vspace{-0.6cm} {\large miniRBO} \\
    \hline
  \end{tabular}
\end{center}
\end{minipage}


\end{document}